\documentclass[accepted]{melba}

\usepackage{mwe} 

\usepackage{amsmath,amsfonts}
\usepackage{amsmath,amssymb,amsfonts}
\usepackage{algorithmic}
\usepackage{graphicx}
\usepackage{textcomp}
\usepackage{booktabs}
\usepackage{xcolor}
\usepackage{siunitx}
\usepackage{subcaption}
\usepackage{caption}
\usepackage{mwe} 
\usepackage{array,multirow,graphicx}
\usepackage{float}

\melbaid{2025:018}  
\doi{https://doi.org/10.59275/j.melba.2025-8d4c}
\melbaauthors{Kenia, Li, Srivastava, and Thakoor}  
\email{rk3291@columbia.edu}
\volume{3}
\firstpageno{402}  
\melbayear{2025}  
\datesubmitted{2025-04-15}  
\datepublished{2025-09-08}  

\ShortHeadings{AI-CNet3D: Anatomically-Informed Cross-Attention Network}{Kenia \emph{et al.}}

\title{AI-CNet3D: An Anatomically-Informed Cross-Attention Network with Multi-Task Consistency Fine-tuning for 3D Glaucoma Classification}

\author{
	\firstname Roshan \surname Kenia\aff{1}\orcid{0009-0001-8384-6786},
	\name Anfei Li\aff{2},
	\name Rishabh Srivastava\aff{1},
	\name Kaveri A. Thakoor\aff{1,2,3,4}\orcid{0000-0001-5589-8151}
}
\affiliations{
	\num 1 \addr Department of Computer Science, Columbia University, New York, NY, USA \\
	\num 2 \addr Department of Ophthalmology, Columbia University Irving Medical Center, New York, NY, USA \\
	\num 3 \addr Department of Biomedical Engineering, Columbia University, New York, NY, USA\\
    \num 4 \addr The Data Science Institute, Columbia University, New York, NY, USA\\
}

\abstract{
	Glaucoma is a progressive eye disease that leads to optic nerve damage, causing irreversible vision loss if left untreated. Optical coherence tomography (OCT) has become a crucial tool for glaucoma diagnosis, offering high-resolution 3D scans of the retina and optic nerve. However, the conventional practice of condensing information from 3D OCT volumes into 2D reports often results in the loss of key structural details. To address this, we propose a novel hybrid deep learning model that integrates cross-attention mechanisms into a 3D convolutional neural network (CNN), enabling the extraction of critical features from the superior and inferior hemiretinas, as well as from the optic nerve head (ONH) and macula, within OCT volumes. We introduce Channel Attention REpresentations (CAREs) to visualize cross-attention outputs and leverage them for consistency-based multi-task fine-tuning, aligning them with Gradient-Weighted Class Activation Maps (Grad-CAMs) from the CNN's final convolutional layer to enhance performance, interpretability, and anatomical coherence. We have named this model AI-CNet3D (AI-`See'-Net3D) to reflect its design as an Anatomically-Informed Cross-attention Network operating on 3D data. By dividing the volume along two axes and applying cross-attention, our model enhances glaucoma classification by capturing asymmetries between the hemiretinal regions while integrating information from the optic nerve head and macula. We validate our approach on two large datasets, showing that it outperforms state-of-the-art attention and convolutional models across all key metrics. Finally, our model is computationally efficient, reducing the parameter count by one-hundred--fold compared to other attention mechanisms while maintaining high diagnostic performance and comparable GFLOPS.
    Our code is available at \href{https://doi.org/10.5281/zenodo.17082118}{10.5281/zenodo.17082118}.
    }


\begin{document}

\twocolumn[\maketitle]

\section{Introduction}
\begin{figure*}[htb]
    \centering
    \begin{subfigure}[t]{0.25\textwidth}
        \centering
        \captionsetup{justification=centering}
        \includegraphics[width=.95\textwidth]{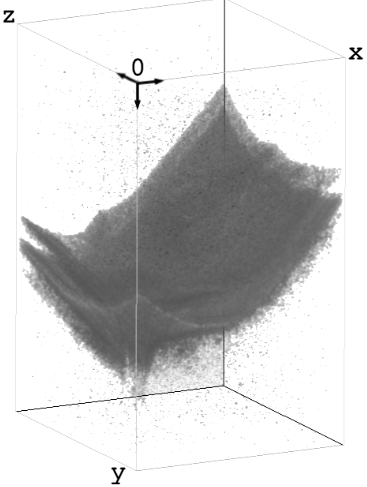}
        \caption{Input Volume 128x192x112}
        \label{fig:res:a}
    \end{subfigure}%
    ~ 
    \centering
    \begin{subfigure}[t]{0.25\textwidth}
        \centering
        \captionsetup{justification=centering}
        \includegraphics[width=.95\textwidth]{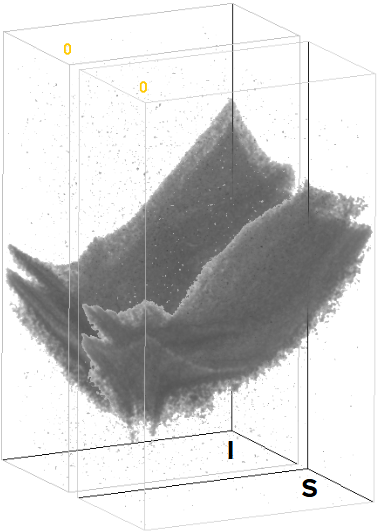}
        \caption{Superior/Inferior Hemiretinas Half-volume 64x192x112}
        \label{fig:res:b}
    \end{subfigure}%
    ~ 
    \centering
    \begin{subfigure}[t]{0.25\textwidth}
        \centering
        \captionsetup{justification=centering}
        \includegraphics[width=.95\textwidth]{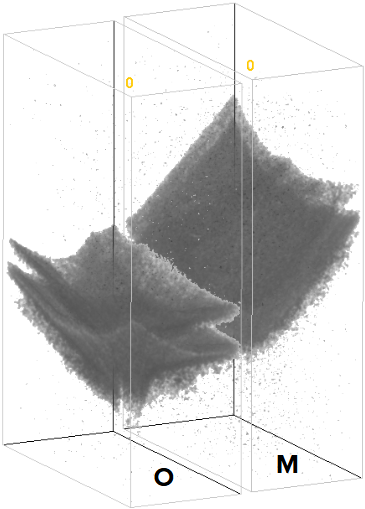}
        \caption{ONH/Macula Split Half-volume 128x192x56}
        \label{fig:res:c}
    \end{subfigure}%
    ~ 
    \centering
    \begin{subfigure}[t]{0.25\textwidth}
        \centering
        \captionsetup{justification=centering}
        \includegraphics[width=.95\textwidth]{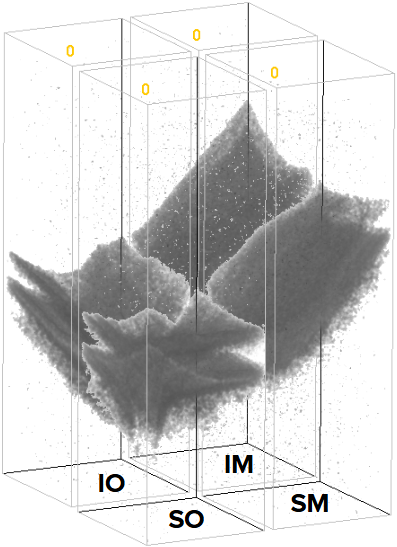}
        \caption{4 Way Widefield Split Quarter-volume 64x192x56}
        \label{fig:res:d}
    \end{subfigure}%
    ~ 
    \caption{An example of how an OCT volume from Dataset 1 can be split along different axes to separate the anatomy in the volume. $CA_H$ is computed using the inferior (I) and superior (S) hemiretinas. $CA_{NA}$ is computed using the ONH (O) and macula (M). $CA_{H-NA}$ is computed using the inferior ONH (IO), inferior macula (IM), superior ONH (SO), and superior macula (SM). }
\label{fig:res}
\end{figure*}

Glaucoma is one of the leading causes of irreversible blindness worldwide \citep{steinmetz2021causes, quigley2006number}. However, as a chronic condition, glaucoma has a slow and gradual onset and often shows no noticeable symptoms in its early stages. Regular eye exams are essential for its classification and treatment. As nerve fiber layer damage is thought to be one of the hallmarks of glaucoma, optical coherence tomography (OCT) has become a widely used tool for glaucoma detection and diagnosis due to its ability to capture high-resolution 3D volumes of the optic nerve and retina \citep{geevarghese2021optical, bussel2014oct}. The raw 3D volumes are then preprocessed and formatted into 2D OCT reports that can facilitate clinical decision-making by ophthalmologists.

Current OCT reports extract the retinal nerve fiber layer (RNFL) and ganglion cell complex (GCC) thicknesses to help detect the presence and degree of glaucomatous damage. Traditionally, a RNFL defect with a spatially-correlated GCC defect is considered to represent an optic neuropathy, and a specific arced projection of this damage (called an arcuate) involving the superior and/or inferior optic nerve is commonly seen in glaucomatous optic neuropathy. Therefore, the ability to detect correlations between RNFL-GCC defects and correctly identify arcuate patterns are crucial for diagnosing glaucoma using OCT. With deep learning, there is potential to both automate this process and deliver near expert-level care to regions with poor access to glaucoma specialists. Furthermore, the traditional 2D OCT report has heavily relied upon superficial features such as the RNFL and GCC \citep{steinmetz2021causes, chen2018spectral}. 3D models now allow for analysis of the entire 3D OCT volume that contains both superficial structures as well as previously unused deeper structures, which may lead to improvement in glaucoma classification over traditional approaches.

Deep learning has rapidly evolved into a powerful technology for automatically extracting features from images, enabling tasks such as detection, classification, and segmentation. While initially applied to 2D natural scene images, its capabilities have expanded to handle 3D volumes and video data, broadening its impact across various domains, from activity recognition to disease diagnosis \citep{tran2015learning, bertasius2021space, 9112644}. For glaucoma diagnosis, \cite{Maetschke_2019} introduced one of the first 3D CNNs capable of classifying raw, unsegmented OCT volumes of the optic nerve head (ONH). They demonstrated the superiority of using a 3D deep learning model over classical feature-based machine learning algorithms. In addition, by computing Class Activation Maps (CAM) \citep{zhou2016learning}, they found the 3D CNN identified regions typically associated with glaucoma such as the neuroretinal rim, optic disc cupping, and the lamina cribrosa.

Convolutions within CNNs are highly effective at extracting local features \citep{dilated}, but when applied to 3D data, their limited receptive fields can pose challenges. Information in 3D volumes is often sparsely distributed or spread over large regions \citep{ye2012sparse}, making it difficult for convolutions alone to capture global context. In contrast, transformer-based methods inherently provide global attention mechanisms, allowing for a more comprehensive understanding of the data. When combined with 3D convolutions, these approaches synergistically capture both local and global information, significantly enhancing feature representation \citep{unetrpp}. However, this advantage afforded by incorporating attention comes with the drawback of significantly increased computational complexity, which is further amplified when dealing with volumetric data.

In medical imaging, obtaining high-quality data is inherently challenging due to limitations in acquisition, cost, and patient variability. Unlike natural image datasets, where large-scale labeled collections are readily available, medical datasets are often small, imbalanced, and difficult to annotate due to the requirement of expert clinical input. This scarcity of annotated data creates a significant barrier to training robust and generalizable deep learning models. Semi-supervised and unsupervised learning techniques have emerged as powerful solutions to mitigate these challenges by leveraging unlabeled data to enhance model performance. Recent contrastive learning approaches, such as SimCLR \citep{chen2020simple} and BYOL \citep{grill2020bootstrap}, encourage representations of augmented views of the same sample to be similar in a shared latent space, reducing reliance on labeled data. In the medical domain, MedCLIP \citep{wang2022medclip} extends contrastive learning by aligning medical images with textual descriptions, capturing richer semantic relationships. 

Extending these capabilities to 3D is even more challenging due to the pronounced scarcity of valuable (labeled or unlabeled) 3D medical data. Authors of SliViT \citep{avram2023slivit} make this leap by leveraging 2D data from 3D OCT volumes via a 2.5D approach that enables robust performance across multiple tasks in three imaging modalities, even with fewer than 700 annotated volumes. Lee and colleagues do this too via their OCTCube approach \citep{liu2024octcubem3dmultimodaloptical}, where a 3D foundation model trained on over 26,000 OCT volumes is extended with contrastive learning to achieve state-of-the-art retinal disease prediction. Similarly, Swin UNETR \citep{tang2022self} integrates multiple self-supervised strategies, such as inpainting, contrastive learning, and rotation correction, to learn more robust feature representations for 3D medical volumes. Our approach goes beyond these past studies by combining 3D OCT volumes, cross-attention mechanisms, and multi-task (supervised plus unsupervised) fine-tuning that enforces visualization consistency to improve model generalization, reduce annotation burdens, and enable AI-driven diagnostic tools that are both data-efficient and clinically reliable.


Volumetric OCT data contains key information for disease diagnosis oriented in a specific manner based on anatomy and pathophysiology. These features can be harnessed to create a more efficient and meaningful attention mechanism. The superior and inferior hemiretinas are the two parts of the retina that are divided by a horizontal line that runs through the fovea and ONH. Within each hemiretina, the ganglion cells send their axonal projections towards the ONH, forming the RNFL. To address the shortfalls of the conventional practice of condensing 3D OCT volume information into 2D reports, which results in the loss of key structural details for glaucoma classification, we propose a novel hybrid deep learning model. This model integrates cross-attention mechanisms into a 3D convolutional neural network (CNN), enabling the extraction of critical features from the superior and inferior hemiretinas, as well as from the optic nerve head (ONH) and macula, within OCT volumes. Additionally, as a fine-tuning and regularization step, we enforce consistency between visualizations from convolutional and attention layers, ensuring alignment between spatial feature extraction and transformer-based representations. Our contributions are as follows:

\begin{itemize}
\item We show the added benefit of our model, an Anatomically-Informed Cross-attention Network operating on 3D data, AI-CNet3D (AI-`See'-Net3D), achieved through the hybrid use of 3D CNNs with cross attention; by dividing the 3D volume and applying cross-attention, our model enhances glaucoma classification by capturing asymmetries between the hemiretinal regions while integrating information from the ONH and macula.

\item We introduce a novel Channel Attention REpresentation (CARE), which provides direct visualization of channel attention outputs, offering a more precise and interpretable alternative to conventional Grad-CAM-based methods.

\item  By enforcing consistency between attention and convolutional visualizations, our model bridges the gap between these complementary feature extraction methods, improving robustness and interpretability.

\item We validate our approach on two datasets (one proprietary and one publicly available), showing that it outperforms state-of-the-art attention models across all key metrics and conduct ablation studies which highlight the optimal positioning of attention within the 3D CNN architecture pipeline.

\item Finally, our model is computationally efficient, reducing the parameter count by one-hundred--fold compared to other attention mechanisms while maintaining high diagnostic performance and comparable GFLOPS.
\end{itemize}

By leveraging anatomical priors, integrating CNNs with cross-attention, and enforcing consistency between feature representations, AI-CNet3D provides a more interpretable, efficient, and clinically relevant approach to 3D OCT analysis.

\section{Related Works}
\subsection{Glaucoma classification}
Glaucoma classification has progressed significantly with the integration of computational methods and deep learning-based approaches. Early work, such as that by  \cite{bock2010glaucoma}, applied appearance-based dimension reduction to color fundus images to develop a glaucoma risk index for classification. The introduction of deep learning marked a pivotal shift, with CNN-based methods becoming a standard for leveraging large datasets to enhance classification accuracy \citep{mehta2021automated, barros2020machine}. \cite{7318462} pioneered one of the first deep CNN architectures specifically for glaucoma classification from fundus images. Later, \cite{li2019attention} extended CNN architectures by incorporating attention maps, enabling more focused analyses of critical retinal regions for improved interpretability. 

Recent advances have shifted towards the utilization of RNFL data from OCT reports, supported by the findings of \cite{hood2022detecting}. \cite{thakoor2020robust} developed a CNN model specifically for RNFL analysis from OCT images, employing concept activation vectors to compare model outputs with clinician eye fixations, adding a layer of clinical relevance. Additionally, \cite{luo2023harvard} introduced a large-scale OCT dataset to support semi-supervised learning, using a generalization-reinforced pseudo-labeling model to improve classification in cases with limited labeled data. 

While the majority of existing research focuses on 2D imaging, recent progress by \cite{Maetschke_2019} and \cite{9112644} has led to the development of CNN approaches for 3D OCT-based glaucoma classification, laying the groundwork for further innovation in 3D imaging modalities. However, these approaches have been validated on OCT volumes from only a single device manufacturer, leaving their generalizability across different imaging systems untested. Ensuring cross-manufacturer robustness is essential for the widespread clinical adoption of such methods, regardless of the OCT device used. We outline an approach for cross-manufacturer training as future work in Appendix \ref{federated_training}.

\subsection{Cross-Attention and 3D Attention}
\label{related3Datt}
In self-attention, the keys and values are derived from the same source as the queries, whereas in cross-attention, the keys and values come from a different source than the queries, allowing the model to focus on external information during processing \citep{vaswani2017attention, lin2022cat}. \cite{chen2021crossvit} introduced CrossViT, that proposes a dual-branch transformer that processes image patches of varying sizes through separate branches, using multiple attention layers to fuse the tokens and enhance image features. To improve computational efficiency, they introduce a cross-attention-based token fusion module, where a single token from each branch serves as a query to exchange information between branches.

Modeling 3D attention is essential for tasks involving volumetric data, such as medical imaging, where spatial relationships extend across three dimensions. By capturing these interactions, 3D attention allows models to learn more complex spatial features with long-range anatomical dependencies, improving the accuracy and robustness of tasks like segmentation or classification \citep{islam2020brain, wang2019volumetric}. \cite{unetrpp} introduced Efficient Paired Attention (EPA), a computationally efficient method for calculating both spatial and channel self-attention in 3D volumes for segmentation tasks by using shared weights. They integrated EPA into a transformer block, utilizing it during both the downsampling and upsampling stages of a convolutional UNet, significantly enhancing performance while reducing computational costs. However, this reduction in computational costs introduced a significant bottleneck, as a large feature volume is condensed into a small vector for spatial attention, creating a substantial constraint. As such, striking a balance between parameter efficiency and anatomical sensitivity remains a key challenge in designing attention modules for 3D medical tasks \citep{xie2023attention, cao2022swin}.

While standard 3D transformers and attention-augmented CNNs apply self-attention or spatial attention across entire volumes in a data-driven manner, our cross-attention mechanism is specifically constrained by retinal anatomy, computing attention only between anatomically meaningful regions (superior-inferior hemiretinas, macula-ONH pairs). Unlike generic attention mechanisms that must learn spatial relationships from scratch like EPA \citep{unetrpp}, our approach embeds established medical knowledge directly into the architecture, enabling more efficient and clinically relevant feature learning.

\subsection{3D Visualization}

\cite{Maetschke_2019} and \cite{9112644} achieved visualization of a 3D CNN model using 3D Grad-CAM, which highlights important regions of the input by backpropagating gradients from the class score to the final convolutional layer, computing the significance of feature maps, and generating a heatmap to identify key areas influencing the model's prediction. While convolutions excel at extracting local features, they may struggle to capture global context effectively, often resulting in sparse Grad-CAM heatmaps. For a clinician, it may be difficult to interpret the model's decision-making process using only 3D Grad-CAMs, as they are only applicable to convolutional layers and thus may not provide insights into the mechanisms of hybrid CNN-attention models. 

Attention rollout was introduced as a post hoc method to trace how information flows from the input layer to the embeddings in higher layers of a transformer by calculating attention across multiple paths between nodes in different layers \citep{abnar2020quantifying}. This is done by recursively multiplying attention weight matrices across layers, allowing for the total amount of information transferred between any two nodes to be computed. \cite{chefer2021transformer} extended attention visualization for computer vision classification tasks by developing a method that applies Deep Taylor Decomposition to assign local relevance scores, which are then propagated through the attention layers to improve interpretability.

There has been limited work on visualizing the components of 3D attention mechanisms. Typically, 3D attention is applied within a single layer, which restricts the use of advanced visualization methods such as attention rollout or \cite{chefer2021transformer}, both of which require multi-layer attention propagation. This lack of suitable visualization tools makes it challenging to interpret and analyze how attention mechanisms function in 3D models, limiting their transparency and usability in clinical applications. 

\subsection{Consistency-based Learning}

Given the scarcity of labeled 3D medical imaging data, regularization techniques have been explored as a means of improving model generalization without relying on large-scale annotations. One particularly effective approach is visualization consistency, which enforces stability in the activations a model produces for a given input. By ensuring that different transformations of the same data yield consistent feature representations, these methods help reduce sensitivity to spurious variations, ultimately enhancing model robustness.

Prior work has predominantly focused on enforcing consistency within convolutional layers using natural images. For example, \cite{guo2019visual} encourage consistency between the class activation maps (CAMs) from the last convolutional layer for both the original and augmented views of the same data. Similarly, \cite{li2020unsupervised} enforce consistency between images that share similar features, while \cite{wang2019sharpen} extend this idea by maintaining Grad-CAM consistency across different CNN layers for a single input. \cite{xu2020multi} further generalize these ideas by enforcing consistency between learned attention maps across augmented images and multiple layers. More recently, \cite{mirzazadeh2023atcon} introduced an alternative approach by enforcing consistency between two different visualization techniques, Grad-CAM and Guided Backpropagation, to improve the quality of generated attention maps.

Despite these advancements, existing visualization consistency techniques remain fundamentally limited in scope, as they are primarily constrained to convolutional layers. With the increasing adoption of hybrid models that integrate convolutional layers with attention mechanisms, there is a pressing need to extend consistency-based learning beyond traditional CNNs. Unlike convolutional models, attention mechanisms dynamically reweight feature importance across an entire image or volume, making them susceptible to different types of instability. By enforcing consistency not only within CNN-based feature maps but also across attention outputs and hybrid representations, we can ensure that the model maintains spatial and anatomical coherence potentially even in low-data/low-labeled-data regimes. In the context of 3D medical imaging, where anatomical structures and pathological regions vary significantly, consistency-based learning offers a promising pathway to improve model interpretability, reduce sensitivity to noise, and enhance generalization across unseen cases.

\subsection{Resource-Efficient Networks}
Reducing parameter count and memory usage in 3D medical imaging models is critical due to the high computational demands imposed by volumetric data, which are often orders of magnitude greater than those of 2D images. While recent approaches have improved model speed during training and inference \citep{unetrpp, pang2023slim, liu2024swin}, they often overlook parameter efficiency, which is equally important for deployment. In portable or point-of-care settings, many applications require real-time inference on devices with limited computational resources \citep{shaker2023swiftformer}. Moreover, large models not only hinder deployment in such environments but also increase the risk of overfitting, especially when trained on small annotated medical datasets \citep{shaikhina2017handling}. To address these challenges, efficient architectures that compress spatial and channel information while preserving critical anatomical context are essential for practical, scalable, and clinically viable solutions.

\begin{figure*}[htb]

\begin{minipage}[b]{1.0\linewidth}
  \centering
  \centerline{\includegraphics[width=\textwidth]{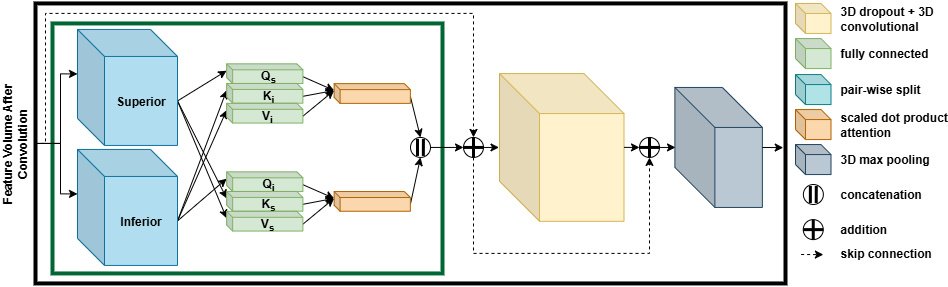}}
\end{minipage}
\caption{Our cross-attention mechanism operates between two pairs of subsections from the feature volume, as highlighted in the {\textbf{\color{OliveGreen}green}} box. In this example, we are computing cross-attention between the superior (S) and inferior (I) hemiretina split used for $CA_{H}$. For $CA_{H-NA}$ (not visualized here), we would repeat the calculation performed in the {\textbf{\color{OliveGreen}green}} box for each pair of quarter-volumes and then concatenate the results before performing the skip connection addition.}
\label{fig:cross-attention}
\end{figure*}


\section{Datasets}
\textbf{Dataset 1} is comprised of 4,932 non-glaucomatous and 272 glaucomatous widefield OCT volumes obtained from Topcon Healthcare, Inc. (Tokyo, Japan) and labeled by OCT experts at Columbia University Irving Medical Center. To create a balanced dataset, we randomly sampled 272 non-glaucomatous volumes during each training trial, resulting in a total of 544 volumes with an equal split of 50\% non-glaucomatous and 50\% glaucomatous cases. This design choice was motivated by early observations that all models tended to overfit to the majority class when trained on imbalanced data even with resampling. For completeness, we provide comparative results of alternative sampling strategies attempted in Appendix \ref{sampling_strategy} and Table \ref{sampling_table}. The original volume dimensions were 128x885x512 (z, y, x) pixels; however, to reduce computational complexity while preserving the y-x aspect ratio, we downsampled the volumes using a uniform scaling affine transformation to 128x192x112 pixels. To ensure a robust evaluation, we divided the dataset into 65\% for training, 15\% for validation, and 20\% for testing, providing a well-balanced selection of samples for model assessment.

\textbf{Dataset 2} is a publicly available dataset provided by \cite{Maetschke_2019} consisting of OCT scans centered on the optic nerve head (ONH), acquired from 624 patients using a Zeiss Cirrus SD-OCT scanner (Jena, Germany). After excluding scans with a signal strength below 7, 1,110 high-quality scans were retained for analysis, with 263 scans labeled as healthy and 847 diagnosed with primary open-angle glaucoma (POAG). We will refer to the healthy as non-glaucomatous and the POAG as glaucomatous. Glaucomatous eyes were defined by the presence of visual field defects, confirmed by at least two consecutive abnormal test results. Just as with Dataset 1, we randomly sampled 263 glaucomatous volumes during each training trial, resulting in a total of 526 volumes with an equal split of 50\% non-glaucomatous and 50\% glaucomatous cases. We utilized the original volume dimensions of 64x128x64 (z, y, x) pixels. The dataset was split into 65\% training, 15\% validation, and 20\% testing, ensuring that scans from the same patient were not split across different sets and allowing more test data for evaluation.

\section{Methods}
\subsection{Anatomically Informed Cross–attention}

\subsubsection{3D Superior-Inferior Cross-attention}
\label{method-cross}

When referring to the retina, the superior hemiretina denotes the nerve fibers originating from the upper portion of the retina, while the inferior hemiretina refers to those arising from the lower part. Together, they represent distinct sections of the retina, each responsible for transmitting visual information from the top and bottom halves of the eye, respectively. In cases of asymmetric glaucoma, either the superior or inferior hemiretina is typically affected. Despite this, no existing glaucoma classification models, to our knowledge, have leveraged this asymmetry to enhance classification. We introduce a novel 3D cross-attention mechanism that leverages the distinct information found in the superior and inferior hemiretinas within an OCT volume to improve glaucoma classification. A standard attention mechanism might struggle to fully capture the nuanced differences between these two regions. To address this, the feature volume can be split along the z-axis, enabling cross-attention between the superior and inferior hemiretinas. This approach allows for relative comparison of the regions (e.g., a healthy inferior hemiretina serves as a reference for a superior hemiretina with disease or vice versa), thereby capturing the asymmetry and enhancing classification capabilities.

In \cite{unetrpp}, an efficient approach to computing spatial and channel attention for volumetric data is presented, leveraging shared key and query weights across the two separate self-attention calculations. For channel attention, the method applies a standard linear projection on the input volume, generating a channel value vector, followed by self-attention. Spatial attention, designed to minimize complexity, first applies a linear projection to form a spatial value vector. This vector is then projected down to a lower dimension, $p$, which is significantly smaller than the number of tokens, $n$. This adjustment reduces the computational complexity from $O(n^2)$ to $O(np)$, yet introduces a notable bottleneck. Specifically, projecting a volume of dimensions $C \times D \times H \times W$ into $C \times p$ entails substantial information loss, impacting the model's ability to capture critical spatial details. Additionally, this second projection layer introduces a high parameter count, as illustrated in Figure \ref{fig:complexity}, further complicating model efficiency and potentially affecting scalability.

Therefore, to increase scalability while still enforcing our anatomical prior, we focus on using only channel attention. We found that projecting the entire spatial dimension of the feature volume into a small vector was not beneficial for model training (ablation studies in Appendix \ref{remove_spatial}). Instead of directly projecting the entire feature volume of size $C \times D \times H \times W$ into query, key, and value vectors for self-attention, we split the input feature volume, $I_v$, along the z-axis into two feature volumes of size $C \times \frac{D}{2} \times H \times W$, representing the superior and inferior hemiretinas as shown in Figure \ref{fig:res:b}. In the first cross-attention step, the superior feature volume is projected into a query vector $Q_s$, while the inferior feature volume forms the key $K_i$ and value $V_i$ vectors. We then apply scaled dot product attention across the channel dimension, computed as:
\begin{equation}
    \label{firstcross}
    A_{SI}(Q_s, K_i, V_i) = \text{softmax}\left(\frac{Q_s K_i^T}{\sqrt{d_k}}\right) V_i
\end{equation}
where $d_k$ is the dimensionality of the key vectors. In the second step, the roles are reversed: the inferior feature volume is projected as the query $Q_i$, while the superior feature volume forms the key $K_s$ and value $V_s$ vectors. Another round of scaled dot product attention is applied:
\begin{equation}
    \label{secondcross}
    A_{IS}(Q_i, K_s, V_s) = \text{softmax}\left(\frac{Q_i K_s^T}{\sqrt{d_k}}\right) V_s
\end{equation}
These two attention outputs are then concatenated together and reshaped to the original input size of $C \times D \times H \times W$ as shown in Fig. \ref{fig:cross-attention}. A skip connection is then used to add the original $I_v$ and results in our hemiretinal cross-attention, $CA_{H}$, as calculated in Equation \ref{ca_h}. This bidirectional attention mechanism allows the model to capture critical interactions between the superior and inferior hemiretinas, improving the representation of the retinal structure. We call this model $\text{AI-CNet3D}_H$, illustrating its combination of information learned between the superior and inferior hemiretinas within the volume.

\begin{equation}
\label{ca_h}
CA_{H} = (A_{SI} \| A_{IS}) \oplus I_v
\end{equation}

\subsubsection{Widefield (4-Way) Cross-attention}
The majority of OCT datasets do not include a widefield view, but Dataset 1 used in this study and described above does. This means that it includes both the optic nerve head (ONH) and the macula within the volumetric scan of the retina. The distinction between ONH and macula is made by splitting the volume along the x-axis. As shown in Figure \ref{fig:res:d}, combining this with our split along the z-axis for the superior and inferior hemiretinas allows us to obtain four subvolumes from the input: the superior ONH ($SO$), superior macula ($SM$), inferior ONH ($IO$), and inferior macula ($IM$). We know from Section \ref{method-cross} that superior and inferior hemiretinas can be used to compute cross-attention. Furthermore, ONH and macular half-volumes are anatomically related, while opposing ONH and macular quarter volumes (e.g., superior ONH and inferior macula) are not anatomically related.

Leveraging this insight, we can enhance our 3D cross-attention mechanism for widefield OCT volumes. In Section \ref{method-cross}, we computed cross-attention between the superior and inferior hemiretinas. With widefield volumes, we extend this to compute cross-attention specifically between anatomically related pairs within $SO$, $SM$, $IO$, and $IM$. Given a quarter-volume, $x$, we project it to a query vector $Q_x$, while the other paired quarter-volume, $y$, forms the key $K_y$ and value $V_y$ vectors. We compute the scaled dot product attention, $A_{xy}$, just as in Equation \ref{firstcross}, and then reverse roles using projected vectors $Q_y$, $K_x$, and $V_x$ for our second scaled dot product attention calculation, $A_{yx}$, just as in Equation \ref{secondcross}. We apply this same sequence of steps to the other two quarter-volumes, $w$ and $z$, and obtain $A_{wz}$ and $A_{zw}$. We can then compute the cross-attention for these two pairs, $xy$ and $wz$, on our volume by concatenating the results as:

\begin{equation}
CA_{xywz} = (A_{xy} \| A_{yx}) \| (A_{wz} \| A_{zw})
\end{equation}

In this case, we compute cross-attention twice. Our first set of quarter volumes $x$, $y$, $z$, and $w$ are represented by $SO$, $SM$, $IO$, and $IM$ respectively and correspond to computing cross-attention, $CA_{SupInf}$ within each hemiretina. Our second set of quarter volumes $x$, $y$, $z$, and $w$ are represented by $SO$, $IO$, $SM$, and $IM$ respectively and correspond to computing cross attention, $CA_{MacONH}$, within each macula and ONH half-volume. These operations are added together using a skip connection along with the original volume, $I_v$, as:

\begin{equation}
CA_{H-NA} = CA_{SupInf} \oplus CA_{MacONH} \oplus I_v
\end{equation}

where $H-NA$ indicates using the hemiretinas and ONH and macula. We call this model $\text{AI-CNet3D}_{H-NA}$, illustrating its  combination of information learned between the superior and inferior hemiretinas within the volume and between the macula and ONH.

\subsubsection{3D Macula-Optic Nerve Head Cross-attention}
For completeness, using our widefield volumes, we follow the same steps used in Section \ref{method-cross}, but split our volume only once along the width axis (x-axis) into macula and optic nerve head (ONH) halves as shown in Figure \ref{fig:res:c}. This enables us to compute the cross-attention between the macula (neurons) and ONH (axons) as $CA_{NA}$, leveraging the distinct information presented in each. We call this model $\text{AI-CNet3D}_{NA}$, illustrating its combination of information learned between the macula and ONH within the volume.

\subsection{3D CNN}
We adopt the 3D CNN as described in \cite{Maetschke_2019} and modify it to include our cross-attention mechanism. The original network consists of five 3D convolutional layers, each with ReLU activation and batch normalization. The layers use filter banks of size 32-32-32-32-32, with filter dimensions of 7-5-3-3-3 and strides of 2-1-1-1-1. After the final convolutional layer, a global average pooling (GAP) layer is applied with a kernel size set to the smallest spatial dimension after downsampling, and a stride matching the final feature map shape, ensuring the entire spatial volume is reduced to a single value per channel. This adaptive configuration ensures spatially aware aggregation of features before classification. This is followed by a dense layer connected to the softmax output. 

After the second and fourth convolutional layers (optimal position verified in Table \ref{tb1}), we introduce a cross-attention-based feature extraction block. The feature vector is first split into halves or quarters, which are processed through one of the three cross-attention mechanisms described previously, while maintaining a skip connection with the full feature vector. Inspired by \cite{unetrpp}, we apply a 3D dropout layer, followed by a 1x1x1 convolution with another skip connection to refine the feature representation. Finally, we employ 3D max pooling to downsample the attention maps, selectively retaining the highest activation values within each pooling region, which correspond to the most significant features. 


\begin{figure*}[htb]

\begin{minipage}[b]{1.0\linewidth}
  \centering
  \centerline{\includegraphics[width=\textwidth]{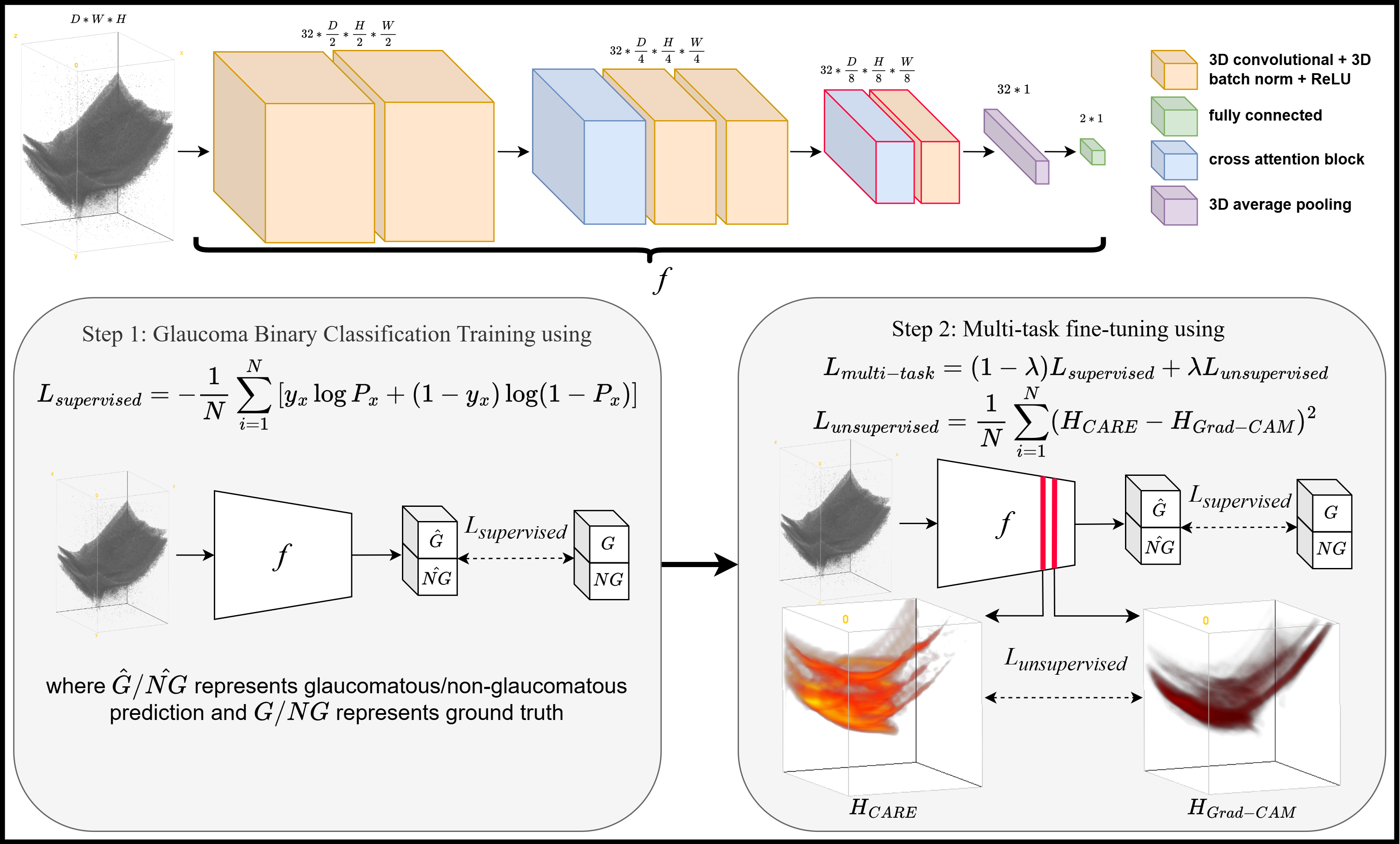}}
\end{minipage}
\caption{Visualization of our AI-CNet3D architecture (with the channel dimension omitted from visualization). We apply multiple layers of convolution along with two cross-attention blocks. Filter banks of size 32 are used consistently across the model. When training with multi-task fine-tuning, we utilize the last cross-attention and convolutional layers highlighted in \textcolor{red}{red} for alignment.}
\label{fig:full-model}
\end{figure*}

\subsection{Channel Attention REpresentation}
\label{CARE}
To complement the 3D Grad-CAM output of the convolutional layers of our network, we introduce a new method called Channel Attention REpresentation (CARE) to visualize the 3D attention layers. This approach translates the attention output, denoted as $CA_0$, into a volumetric representation aligned with the original input volume, thereby enabling interpretable visualization of critical features. The cross-attention output $CA_0$ is obtained post max-pooling (see Fig. \ref{fig:cross-attention}), yielding the layer’s most influential attention features. From this, we specifically isolate the channel dimension to capture attention features from multiple perspectives within the same spatial location. To condense this channel information, we compute the mean attention map by averaging $CA_0$ across the channel dimension $C$:


\begin{equation}
	CA_1 = \frac{1}{C}\sum_{c'=1}^{C} CA_0(c', d, h, w)
\end{equation}

The resulting $CA_1$ reduces dimensional complexity while preserving channel-wise aggregated attention weights. Next, to focus on features with positive contributions, analogous to Grad-CAM, we apply a rectified linear unit (ReLU) activation function to $CA_1$, ensuring that only positive activations contribute to the visualization. For interpretability, the rectified attention map is max normalized to scale its values between 0 and 1, facilitating its use as a heatmap: 

\begin{equation}
   H_{CARE} = \frac{ReLU(CA_1)}{max(ReLU(CA_1)) + \epsilon}
\end{equation}

where $ \epsilon $ is a small constant to prevent division by zero. This simple computation is performed at the last attention layer within the network and can be used with any of the mechanisms explained above. Since it is computed using a deeper network layer, the resulting $D \times H \times W$ dimensions may be smaller than those of the original volume. To address this, we apply 3D interpolation to rescale the $H_{CARE}$ to match the input’s original size only for visualizations (see Fig. \ref{fig:heatmap}), allowing us to overlay it and generate a clear, interpretable heatmap. We ensured that this operation was fully differentiable so that it could be utilized to train our model.

\subsection{3D Grad-CAM}

3D Gradient Weighted Class Activation Maps (3D Grad-CAMs) extend the original 2D Grad-CAM method \citep{selvaraju2020grad} to identify influential regions within a 3D volume, highlighting areas that significantly contribute to the model's class prediction. The gradient of the score $y^c$ for class $ c \in \{0,1\} $ is computed with respect to a 3D feature map $A^k$ at each voxel $ (i, j, d) $. This gradient is globally averaged over the spatial dimensions to compute class-specific weights $ a_{k}^{c} $ for each feature map $ k $:

\begin{equation} 
a_{k}^{c} = \frac{1}{Z} \sum_{i}\sum_{j}\sum_{d}\frac{\partial y^{c}} {\partial A^{k}_{i,j,d}} 
\label{gradCamDef3D} 
\end{equation}

where $ Z $ is a normalization constant representing the total number of voxels. These weights are then used to scale the feature maps, emphasizing the most influential regions. The final 3D Grad-CAM heatmap $H_{Grad-CAM}$ is generated by summing the weighted feature maps across all channels and applying a rectified linear unit (ReLU) to ensure non-negative values. To normalize the heatmap, each voxel intensity is divided by the maximum value across the entire volume:

\begin{equation} 
H_{Grad-CAM} = \frac{ReLU\left(\sum_{k} a_{k}^{c} A^k\right)}{max(ReLU\left(\sum_{k} a_{k}^{c} A^k\right)) + \epsilon}
\label{gradCamDef3D3} 
\end{equation}

where $ \epsilon $ is a small constant to prevent division by zero. This process ensures that the heatmap intensities are scaled between 0 and 1, making them in a visualizable range. The resulting heatmap highlights the regions within the 3D volume that contribute most strongly to predicting glaucoma presence, with warmer colors indicating greater influence on the model’s decision (yellow represents highest influence). 

\subsection{Multi-task Fine-tuning to Enforce Visualization-Based Consistency}

We now present our method to enforce consistency between the attention and convolutional layers of our network. It is known from \citep{selvaraju2020grad} that the last convolutional layer of a CNN captures the most class-discriminative properties compared to earlier layers and therefore is best to use for visualization. We follow this trend and obtain the last convolutional layer's output and the last attention layer output from our model. These outputs are of a much smaller dimension compared to the original input volume (due to downsampling for the convolutional layer and max-pooling for the attention layer), and thus they contain the most relevant information the model uses for classification.

Our goal is to enforce consistency between the hidden features extracted from the cross-attention module and the final convolutional layer, ensuring that features learned by one can be effectively shared with the other. Previous consistency-based loss methods have utilized Pearson correlation, Structural Similarity Index (SSIM), Kullback-Leibler (KL) divergence, and Mean Squared Error (MSE) to compare features. In practice, we found that MSE loss worked best (ablation studies in Appendix \ref{loss_choice}) to enforce feature alignment and improve model performance.  

For a given input $ x $, let $H_{CARE}$ denote the final cross-attention hidden feature heatmap and $H_{Grad-CAM}$ represent the final convolutional layer hidden feature heatmap. We define the unsupervised consistency loss as:  

\begin{equation}
    {L}_{unsupervised} = \frac{1}{N} \sum_{i=1}^{N} (H_{CARE} - H_{Grad-CAM})^2
\end{equation}

Additionally, we compute the binary cross-entropy (BCE) loss between the model's predicted glaucoma probability $ P_x $ and the ground truth $ y_x $:  

\begin{equation}
    {L}_{supervised} = -\frac{1}{N} \sum_{i=1}^{N} \left[ y_x \log P_x + (1 - y_x) \log (1 - P_x) \right]
    \label{BCEloss}
\end{equation}

We found that utilizing only the unsupervised loss while fine-tuning (ablation studies in Appendix \ref{lambda_choice}) leads to model degeneration in terms of classification performance. To jointly optimize for accurate classification and feature consistency, we combine these losses into a multi-task objective function:  

\begin{equation}
    {L}_{multi-task} = (1-\lambda) {L}_{supervised} + \lambda {L}_{unsupervised}
    \label{joint_eq}
\end{equation}

where $ \lambda $ is a weighting factor that controls the influence of the unsupervised consistency loss. This combined loss encourages the network to learn from labeled data while also enforcing consistency between the cross-attention module and the final convolutional layer without any labels, leading to improved feature robustness and interpretability.




\begin{table}\centering
\scriptsize
\caption{Results of ablation study with $\text{AI-CNet3D}$ to determine the optimal placement of cross-attention blocks within the 3D CNN over three trials. Integrating cross-attention after the initial convolutions and before the final convolution yielded the best performance.}
\begin{tabular}{@{}ccc@{}}\toprule
Cross-attention Placement & Avg. Acc. $\pm$ Std. &
  Avg. AUROC $\pm$ Std. \\ \hline
After conv 1 \& 2 &0.7982 ± 0.0327& 0.7980 ± 0.0328 \\ 
After conv 1 \& 3 &0.7951 ± 0.0216& 0.7945 ± 0.022 \\ 
After conv 2 \& 3 &0.7859 ± 0.0312& 0.7856 ± 0.0294 \\ 
After conv 2 \& 4 &\textbf{0.8165 ± 0.0375}& \textbf{0.8174 ± 0.0353} \\ 
After conv 2 \& 5 &0.7584 ± 0.0189& 0.7602 ± 0.0154 \\ 
After conv 3 \& 4 &0.7706 ± 0.0259& 0.7717 ± 0.0246 \\ 
After conv 3 \& 5 &0.8073 ± 0.0417& 0.8072 ± 0.0405 \\ 
\bottomrule
\end{tabular}
\captionsetup{font=small}
\label{tb1}
\end{table}

\begin{table*}\centering
\caption{Performance evaluation of the baseline 3D CNN, the hybrid EPA CNN model, TimeSformer, ViT, SE-ResNeXt, M3T, Med3D, and our proposed hybrid cross-attention CNN models on Dataset 1 (Topcon) and Dataset 2 (Zeiss). We report p-values for Mann Whitney U tests with our $\text{AI-CNet3D}_{H}$ model
in parentheses after the standard deviations.}
\resizebox{\textwidth}{!}{%
\begin{tabular}{@{}cllllll@{}}\toprule
& Model Type &
  Avg. Test Acc. $\pm$ Std. &
  Avg. Test Spec. $\pm$ Std. &
  Avg. Test Sens. $\pm$ Std. &
  Avg. Test AUROC $\pm$ Std. &
  Avg. Test F1 Score $\pm$ Std. \\ \midrule
\multirow{8}{*}{\rotatebox[origin=c]{90}{\large\textbf{Topcon}}}

& ViT \citep{vit}& 
0.6422 ± 0.0682 (0.008) & 
0.7241 ± 0.1803 (0.151) & 
0.5556 ± 0.2843 (0.032) & 
0.6398 ± 0.0772 (0.008) & 
0.5503 ± 0.2755 (0.008) \\

& TimeSformer \citep{bertasius2021space}& 
0.7670 ± 0.0336 (0.094) & 
0.7726 ± 0.0639 (0.421) & 
0.7614 ± 0.0729 (0.151) & 
0.7670 ± 0.0313 (0.095) & 
0.7693 ± 0.0300 (0.032) \\ 

& SEResNeXt50 \citep{squeezeexcite}& 
0.7908 ± 0.0220 (0.248) & 
0.8264 ± 0.0246 (0.841) & 
0.7571 ± 0.0451 (0.151) & 
0.7917 ± 0.0193 (0.222) & 
0.7871 ± 0.0208 (0.209) \\ 

& M3T \citep{jang2022m3t} & 
    0.6532 ± 0.0927 (0.016) &
    0.5942 ± 0.2967 (0.056) &
    0.7143 ± 0.1492 (0.151) &
    0.6542 ± 0.0919 (0.016) &
    0.6785 ± 0.0585 (0.008) \\ 
    
& Med3D \citep{chen2019med3d} & 
    0.6220 ± 0.0825 (0.008) &
    0.4282 ± 0.2220 (0.008) &
    \textbf{0.8179 ± 0.1288} (0.691) &
    0.6230 ± 0.0728 (0.008) &
    0.6886 ± 0.0452 (0.016) \\ 
    
& Base 3D CNN \citep{Maetschke_2019} &  
    0.6073 ± 0.0875 (0.008) &
    0.6062 ± 0.3395 (0.421) &
    0.6122 ± 0.3539 (1.000) &
    0.6092 ± 0.0907 (0.008) &
    0.5453 ± 0.2526 (0.008) \\

& EPA \citep{unetrpp} &
0.7872 ± 0.0404 (0.346) &
0.8235 ± 0.0372 (0.917) &
0.7539 ± 0.0593 (0.310) &
0.7887 ± 0.0385 (0.421) &
0.7831 ± 0.0429 (0.463) \\

& $\text{AI-CNet3D}_{NA}$ (Ours) & 
0.8037 ± 0.0356 & 
0.7992 ± 0.0669 & 
0.8076 ± 0.0253 & 
0.8034 ± 0.0373 & 
0.8090 ± 0.0249\\ 

& $\text{AI-CNet3D}_{H-NA}$ (Ours) & 
0.7945 ± 0.0356 & 
0.7936 ± 0.0820 & 
0.7921 ± 0.0457 & 
0.7928 ± 0.0359 & 
0.7987 ± 0.0233\\ 

& $\text{AI-CNet3D}_{H}$ (Ours) &  
\textbf{0.8183 ± 0.0340} & 
\textbf{0.8290 ± 0.0604} & 
0.8063 ± 0.0203  & 
\textbf{0.8176 ± 0.0339} & 
\textbf{0.8204 ± 0.0288} \\  
\bottomrule
\multirow{6}{*}{\rotatebox[origin=c]{90}{\large\textbf{Zeiss}}}

    & ViT \citep{vit} & 
    0.7520 ± 0.1167 (0.222) &
    0.8077 ± 0.1190 (0.600) &
    0.6797 ± 0.3416 (1.000) &
    0.7437 ± 0.1269 (0.151) &
    0.6531 ± 0.3276 (0.421) \\

& TimeSformer \citep{bertasius2021space} & 
    0.8179 ± 0.0479 (0.691) & 
    0.8440 ± 0.0714 (0.463) &
    0.7963 ± 0.1073 (0.600) &
    0.8202 ± 0.0455 (0.548) &
    0.8102 ± 0.0545 (1.000) \\ 

& SEResNeXt50 \citep{squeezeexcite} & 
    0.8143 ± 0.0411 (0.691) &
    0.8611 ± 0.0830 (0.548) &
    0.7794 ± 0.1009 (0.173) &
    0.8203 ± 0.0398 (0.841) &
    0.8048 ± 0.0414 (0.421) \\ 

& M3T \citep{jang2022m3t} & 
    0.7920 ± 0.0124 (0.008) &
    0.8561 ± 0.0601 (0.310) &
    0.7295 ± 0.0516 (0.016) &
    0.7928 ± 0.0078 (0.008) &
    0.7756 ± 0.0117 (0.008) \\ 

& Med3D \citep{chen2019med3d} & 
    0.8183 ± 0.0249 (0.841) &
    0.8694 ± 0.0572 (0.346) &
    0.7692 ± 0.0887 (0.310) &
    0.8193 ± 0.0250 (0.548) &
    0.8054 ± 0.0337 (0.310) \\ 

& Base 3D CNN \citep{Maetschke_2019} &  
    0.8300 ± 0.0359 (0.600) &
    \textbf{0.8730 ± 0.0896} (0.463) &
    0.7891 ± 0.0695 (0.346) &
    0.8310 ± 0.0311 (0.310) &
    0.8222 ± 0.0280 (0.173) \\

& EPA \citep{unetrpp} &
    0.8162 ± 0.0429 (0.691) &
    0.8379 ± 0.0753 (0.917) &
    0.8038 ± 0.0810 (0.151) &
    0.8209 ± 0.0438 (0.841) &
    0.8122 ± 0.0398 (0.421) \\

& $\text{AI-CNet3D}_{H}$ (Ours) &  
    \textbf{0.8315 ± 0.0105} & 
    0.8246 ± 0.0336 & 
    \textbf{0.8425 ± 0.0388} & 
    \textbf{0.8336 ± 0.0130} & 
    \textbf{0.8311 ± 0.0134} \\ 
\bottomrule
\end{tabular}
}
\label{tb2}
\end{table*}

\subsection{Model Training and Data Augmentation}
\label{augmentations}

For both Dataset 1 and Dataset 2, the data is randomly shuffled and divided into training, validation, and test sets. Our model is trained for 250 epochs using a batch size of 4, a learning rate of 0.0001, and an early stopping patience of 25 epochs. We use the NAdam optimizer and Binary Cross-Entropy (BCE) loss highlighted in Equation \ref{BCEloss}. Each experiment conducted, for our model and the baseline models, is repeated five times, with the results for accuracy, specificity, sensitivity, AUROC, and F1-Score averaged across these runs.

Once we train our model, it can produce clear and meaningful Grad-CAMs and CAREs. We then fine-tune it for another 250 epochs using the multi-task unsupervised and supervised loss in Equation \ref{joint_eq} to make these two visualizations consistent. We observed that if we attempted joint training from scratch, early incorrect visualizations were learned and propagated through the network for the rest of training. We utilized the same hyperparameters and data splits as our regular BCE training. We found through a hyperparameter search (ablation studies in Appendix \ref{lambda_choice}) that $\lambda = 0.75 \text{ and } 0.5$ lead to the best results for Dataset 1 and 2, respectively.

During each epoch of model training and fine-tuning, we implement standard random grayscale augmentation and rescale the volumes by up to 1.25x. A widefield OCT scan contains information about both the macula and the optic nerve head (ONH) in the retina. To leverage the symmetry between the right and left eyes in relation to the macula and nerve, we apply random reflections across the yz-plane. Additionally, to enhance the model's robustness to variations in the superior and inferior hemiretinas, we also apply random reflections across the xy-plane.
\section{Results}

\setlength\extrarowheight{2mm} 

To optimize attention placement, we conducted an ablation study by positioning our superior inferior cross-attention mechanism after various convolutional layers within the network. As shown in Table \ref{tb3}, the best performance occurred when cross-attention was applied after the second and fourth convolutions. This configuration leverages the convolutional layers' role in feature extraction, with early layers capturing small, local patterns like edges and circles, and later layers integrating these into more complex structures \citep{zeiler2014visualizing}.

We benchmark against the TimeSFormer model from \cite{bertasius2021space}, originally designed for video classification but adapted here for 3D volumes using a joint space-time self-attention mechanism. For a simpler self-attention baseline, we compare against a ViT \citep{vit} adapted for 3D volumes. To evaluate against a strong convolutional architecture, we use SEResNeXt \citep{squeezeexcite}, which integrates Squeeze-and-Excitation blocks within a ResNeXt framework. We also compare our approach with the baseline 3D CNN from \cite{Maetschke_2019}. We further compare against M3T \citep{jang2022m3t}, a multi-plane and multi-slice transformer network that combines 3D CNN, 2D CNN, and state-of-the-art 3D transformer architectures to leverage both local inductive biases and global attention relationships across axial, coronal, and sagittal planes for classification. We lastly evaluate against Med3D \citep{chen2019med3d} with ResNet34 backbone, a heterogeneous 3D network pre-trained on the diverse 3DSeg-8 dataset that demonstrates superior transfer learning capabilities for 3D medical imaging tasks compared to models pre-trained on natural image datasets.

For comparison with a standard spatial and channel attention methods we replace our 3D cross-attention mechanism with the Efficient Paired Attention (EPA) algorithm from \cite{unetrpp}. Since this architecture is also a hybrid CNN-attention method, we can follow the same steps detailed in Section \ref{CARE} to compute attention representations for the joint spatial and channel attention mechanisms used in EPA. These representations can then be used to perform consistency-based fine-tuning, allowing us to compare performance against our $\text{AI-CNet3D}$ models in Table \ref{tb3}.

\subsection{Anatomically Informed Cross–attention}

In Table \ref{tb2}, we compare our 3D channel-wise cross-attention approaches to other attention approaches along with the base CNN presented in \cite{Maetschke_2019}. Examining Table \ref{tb2} performance on our widefield Topcon Dataset 1, we can see our $\text{AI-CNet3D}_{H-NA}$, $\text{AI-CNet3D}_H$, and $\text{AI-CNet3D}_{NA}$ models perform comparably or better than other non-cross-attention baseline approaches across all metrics. Our $\text{AI-CNet3D}_H$ model performs comparably or significantly better on all metrics (except sensitivity) than all other models on Dataset 1, including the specialized medical models M3T and Med3D, which achieve lower performance across most metrics despite being designed for medical imaging tasks. While Med3D achieves a slightly higher sensitivity, it exhibits much greater variability, and notably, our $\text{AI-CNet3D}_H$ model surpasses this performance when fine-tuned, as demonstrated in Table \ref{tb3}.

Our $\text{AI-CNet3D}_H$ cross-attention mechanism consistently outperforms other methods across all metrics other than specificity on our ONH-only Zeiss Dataset 2. (Note: Since Dataset 2 contains ONH-only OCT scans without the macula, $\text{AI-CNet3D}_{H-NA}$ and $\text{AI-CNet3D}_{NA}$ cannot be evaluated.) The Base 3D CNN, SEResNeXt50, EPA, TimeSformer, and the medical-specific models M3T and Med3D demonstrate high specificity on Dataset 2, likely due to learning more conservative decision boundaries from overfitting to simpler non-glaucomatous cases, although often at the cost of lower sensitivity (the Base 3D CNN's high specificity here can also be attributed to the fact that this model was optimized for Dataset 2 volumes). In contrast, our cross-attention approach achieves comparable specificity (and surpasses the other 4 models when fine-tuned, as shown in Table \ref{tb3}) while also improving sensitivity. This indicates a superior ability to reduce both false positives and false negatives. Striking this balance is critical in medical applications, where accurately detecting disease (sensitivity) must be balanced with minimizing false alarms (specificity) to ensure reliable diagnostics. 

\begin{figure*}[htb]

\begin{minipage}[b]{1.0\linewidth}
  \centering
  \centerline{\includegraphics[width=0.65\textwidth]{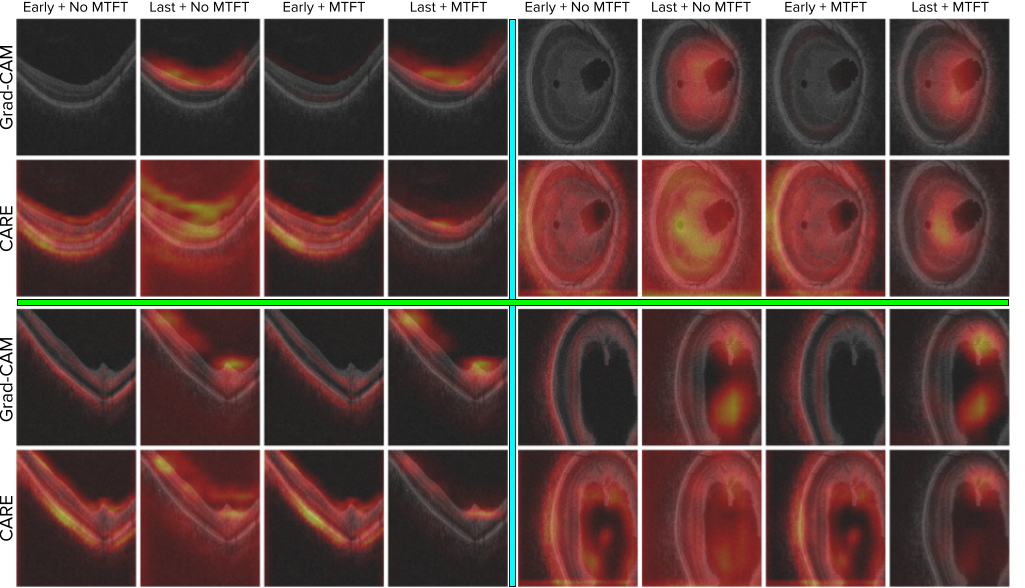}}
  \centerline{\includegraphics[width=0.65\textwidth]{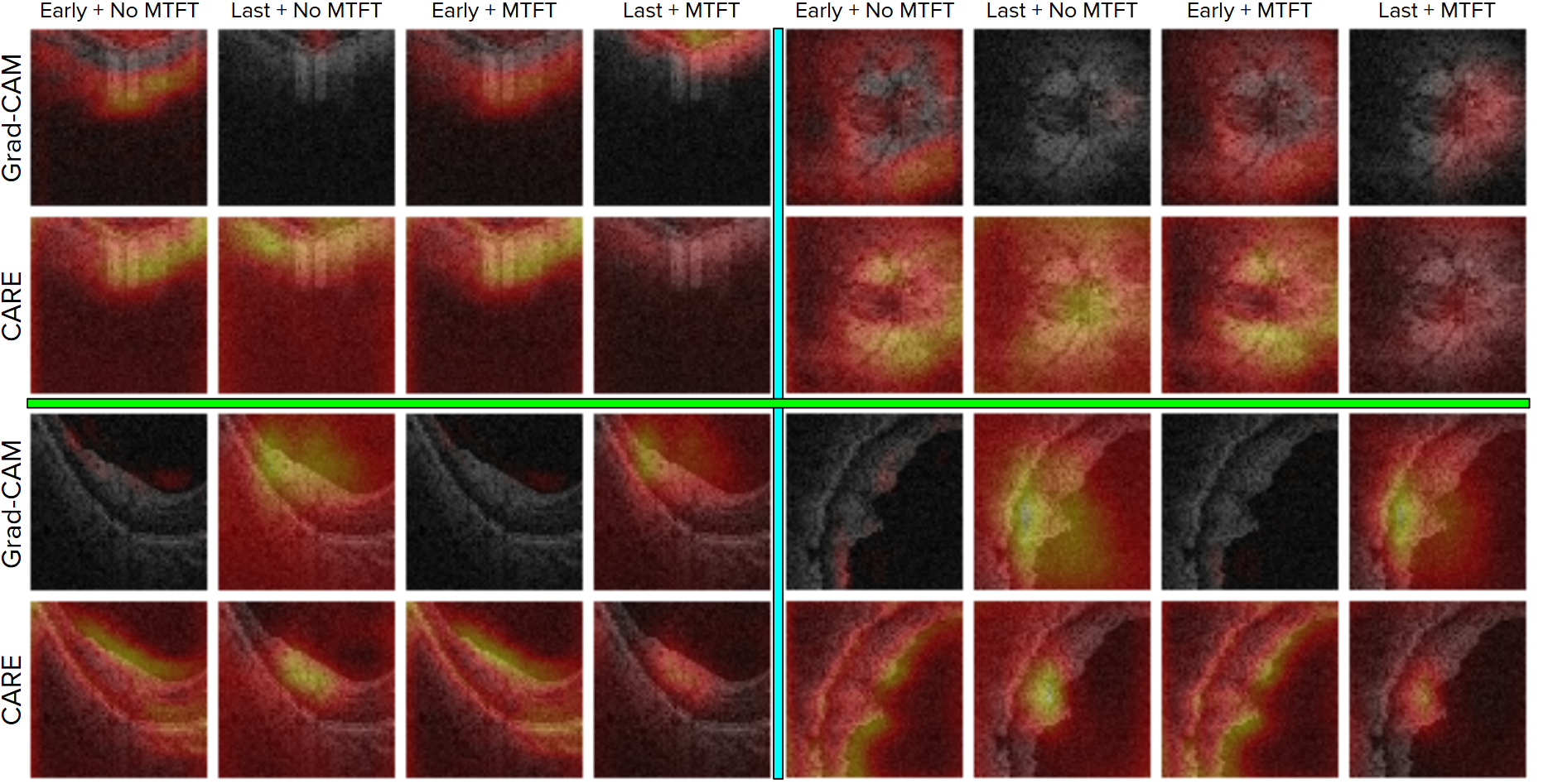}}
\end{minipage}
\caption{\small Comparison of CARE and Grad-CAM visualizations from our top-performing \textbf{AI-CNet3D} model before and after multi-task fine-tuning (MTFT) on \textbf{Dataset 1 (Topcon)} for the first four rows and \textbf{Dataset 2 (Zeiss)} for the last four rows. The heatmaps use a scale of intensities to represent importance, with \colorbox{black}{\textcolor{Goldenrod}{yellow}} indicating the highest relevance, \textcolor{red}{red} indicating moderate importance, and the absence of activation representing the least relevance. The \textbf{No MTFT} model was trained with standard BCE loss for 250 epochs, while the \textbf{MTFT} model was fine-tuned for an additional 250 epochs using a combination of unsupervised MSE loss and supervised BCE loss. \textbf{Early} indicates attention from the second convolutional and first attention layer, whereas \textbf{Last} corresponds to the final convolutional and attention layers. \textbf{True positive} examples are shown above the {\textbf{\color{Green}green}} line, and \textbf{true negatives} are displayed below. \textbf{Axial slices} appear to the left of the {\textbf{\color{Cyan}cyan}} line, while \textbf{coronal slices} are to the right. After MTFT, attention maps from the \textbf{Last} layers exhibit greater consistency, highlighting more stable and interpretable representations.
}
\label{fig:heatmap}
\end{figure*}

\subsection{Multi-task Fine-tuning to Enforce Visualization-Based Consistency}

\begin{table*}\centering
\caption{Performance evaluation of the hybrid EPA-CNN model and our proposed hybrid cross-attention CNN models on Dataset 1 (Topcon) and Dataset 2 (Zeiss), with and without multi-task fine-tuning. The unsupervised weighting in Equation \ref{joint_eq} is set to $\lambda=0.75$ for Dataset 1 and $\lambda=0.5$ for Dataset 2. We report p-values for Mann Whitney U tests between models trained without and with fine-tuning in parentheses after the standard deviations.}

\resizebox{\textwidth}{!}{%
\begin{tabular}{@{}cllllll@{}}\toprule
& Model Type &
  Avg. Test Acc. $\pm$ Std. &
  Avg. Test Spec. $\pm$ Std. &
  Avg. Test Sens. $\pm$ Std. &
  Avg. Test AUROC $\pm$ Std. &
  Avg. Test F1 Score $\pm$ Std. \\ \midrule
\multirow{8}{*}{\rotatebox[origin=c]{90}{\large\textbf{Topcon}}}
& EPA \citep{unetrpp} &
0.7872 ± 0.0404 (0.461) &
0.8235 ± 0.0372 (0.600) &
0.7539 ± 0.0593 (0.834) &
0.7887 ± 0.0385 (0.548) &
0.7831 ± 0.0429 (0.841) \\
& EPA \citep{unetrpp} w/ fine-tuning &
    0.7963 ± 0.0242  & 
    0.8187 ± 0.0869  &
    0.7738 ± 0.0703  &
    0.7962 ± 0.0239  &
    0.7948 ± 0.0256  \\
& $\text{AI-CNet3D}_{NA}$ (Ours) & 0.8037 ± 0.0356 (0.462)
 & 0.7992 ± 0.0669 (0.548)
 & 0.8076 ± 0.0253 (0.674)
 & 0.8034 ± 0.0373 (0.548)
 & 0.8090 ± 0.0249 (0.421)\\ 
 & $\text{AI-CNet3D}_{NA}$ (Ours) w/ fine-tuning & 0.8128 ± 0.0263 
 & 0.8205 ± 0.0535
 & 0.8029 ± 0.0191
 & 0.8117 ± 0.0274
 & 0.8152 ± 0.0144\\ 
& $\text{AI-CNet3D}_{H-NA}$ (Ours) & 0.7945 ± 0.0356 (0.246)
 & 0.7936 ± 0.0820 (0.310)
 & 0.7921 ± 0.0457 (0.917)
 & 0.7928 ± 0.0359 (0.310)
 & 0.7987 ± 0.0233 (0.310)\\
 & $\text{AI-CNet3D}_{H-NA}$ (Ours) w/ fine-tuning & 0.8220 ± 0.0250
 & 0.8659 ± 0.0765
 & 0.7772 ± 0.0646
 & 0.8216 ± 0.0239
 & 0.8170 ± 0.0204\\
& $\text{AI-CNet3D}_{H}$ (Ours) &  0.8183 ± 0.0340 (0.462) & 0.8290 ± 0.0604 (0.917) & 0.8063 ± 0.0203 (0.293)  & 0.8176 ± 0.0339 (0.548) & 0.8204 ± 0.0288 (0.421) \\ 
& $\text{AI-CNet3D}_{H}$ (Ours) w/ fine-tuning & \textbf{0.8312 ± 0.0137} & \textbf{0.8468 ± 0.0482} & \textbf{0.8181 ± 0.0272} & \textbf{0.8325 ± 0.0135}  & \textbf{0.8323 ± 0.0092} \\ 
\bottomrule
\multirow{4}{*}{\rotatebox[origin=c]{90}{\large\textbf{Zeiss}}}

& EPA \citep{unetrpp} &
    0.8162 ± 0.0429 (1.000) &
    0.8379 ± 0.0753 (0.674) &
    0.8038 ± 0.0810 (0.691) &
    0.8209 ± 0.0438 (0.841) &
    0.8122 ± 0.0398 (1.000) \\
& EPA \citep{unetrpp} w/ fine-tuning &
    0.8252 ± 0.0406 &
    0.8666 ± 0.0514 &
    0.7922 ± 0.0977 &
    0.8294 ± 0.0426 &
    0.8163 ± 0.0434 \\
& $\text{AI-CNet3D}_{H}$ (Ours) &  
    0.8315 ± 0.0105  (0.095)& 
    0.8246 ± 0.0336 (0.047) & 
    \textbf{0.8425 ± 0.0388} (1.000)& 
    0.8336 ± 0.0130 (0.095) & 
    0.8311 ± 0.0134 (0.222) \\ 
& $\text{AI-CNet3D}_{H}$ (Ours) w/ fine-tuning &  
    \textbf{0.8573 ± 0.0252} & 
    \textbf{0.8771 ± 0.0237} & 
    0.8392 ± 0.0503 & 
    \textbf{0.8582 ± 0.0249} & 
    \textbf{0.8534 ± 0.0228} \\ 
\bottomrule
\end{tabular}
}
\label{tb3}
\end{table*}

Integrating the cross-attention mechanism into the CNN framework yields a performance improvement. In addition, fine-tuning the model with a combination of unsupervised visualization consistency and BCE losses further enhances its effectiveness. As shown in Table \ref{tb3}, we compare our models against a hybrid EPA CNN model both before and after fine-tuning. Across both datasets, fine-tuning consistently improves generalization for all models, with our $\text{AI-CNet3D}_{H}$ approach achieving the highest overall performance. These results show even greater performance compared to those in Table \ref{tb2} (F1-score of $\text{AI-CNet3D}_{H}$ with fine-tuning is significantly better than that of EPA with fine-tuning, p = 0.03, on Dataset 1). Furthermore, we can see multi-task fine-tuning models applied to Dataset 2 (Zeiss) exhibit improved specificity compared to their non-fine-tuned counterparts, while maintaining sensitivity within each respective model pair. Due to the lower resolution of Dataset 2, the model may struggle to capture anatomical features effectively before fine-tuning. However, after fine-tuning and enforcing layer-wise consistency, it is better able to capture these features, leading to more significant improvements compared to Dataset 1. This shows that the combination of our 3D cross-attention mechanism with consistency enforced between CARE attention representations and Grad-CAMs from convolutional layers within our model helps in precise feature extraction and increased regularization.

\subsection{3D Visualization}

When using CARE and 3D Grad-CAM with our model, we found that extracting visualizations from earlier layers yielded the most aesthetically-pleasing results, as these layers retained the volume's features at a higher resolution most effectively. In Fig. \ref{fig:heatmap}, we compare the CARE from the first cross-attention block with the 3D Grad-CAM from the second convolutional block, using the same model ($\text{AI-CNet3D}_{H}$) since it was trained on both datasets. Solely for visualization purposes, we apply one round of dilation to the Grad-CAM outputs to enhance under-highlighted regions and one round of erosion to the CARE outputs to refine over-highlighted areas. We can see similar regions highlighted by both methods, but CARE extracts information from deeper layers within the retina which we discuss in Section \ref{vizdisc}. We can also see that after fine-tuning, the CARE and Grad-CAM visualizations are more consistent in the last layer since we only utilize deeper layers for alignment training. This means that earlier layers can still capture information using their unique advantages: convolutions for local feature extraction and attention for capturing long-range dependencies. Then, in the deeper parts of the network, these layers synergistically share information learned to make a prediction.

\section{Discussion}

Our targeted approach offers several key advantages over whole-volume spatial self-attention: (1) it directly models known pathophysiological relationships in glaucoma, where asymmetric damage between superior and inferior regions is clinically significant, (2) it dramatically reduces computational complexity by eliminating the need for costly volume projections required in spatial attention (achieving 241k-291k parameters versus 63-88M in standard transformers), and (3) it provides inherent interpretability aligned with clinical understanding. Unlike sequence-based approaches such as State Space Models \citep{gu2021efficiently} or Mamba \citep{gu2023mamba} that process data linearly, our method explicitly captures the spatial relationships critical for understanding retinal pathology. Our CARE visualization reveals that the model leverages deeper retinal structures beyond conventional RNFL analysis, demonstrating how anatomically-constrained attention can uncover clinically relevant features that generic spatial attention might miss.

\subsection{Anatomically Informed Cross–attention}

The introduction of our 3D cross-attention mechanism provides new insights into capturing anatomically-informed structural variations in volumetric OCT data to enhance glaucoma classification performance. The $\text{AI-CNet3D}_{H-NA}$, $\text{AI-CNet3D}_H$, and $\text{AI-CNet3D}_{NA}$ rows in Table \ref{tb2} demonstrate that applying cross-attention yielded consistently better performance compared to models without cross-attention. The improvement seen in our AI-CNet3D models is grounded in current understanding of glaucoma and its pathophysiology. Given that damage to the ganglion cells (GCL) and their axons (RNFL) should be geographically correlated based on anatomy, cross-attention between the macula (where ganglion cell bodies reside) and the optic nerve (where their axons pass) is effective in isolating this correlation over models that do not incorporate cross-attention. Even more pronounced, the anatomical separation of superior and inferior hemiretina allows the $\text{AI-CNet3D}_H$ model to detect early asymmetries through internal control and perform better than its other anatomically-informed variants.

Our results also highlight the versatility and robustness of AI-CNet3D, confirming its generalizability across both ONH-only and widefield OCT imaging formats. Table \ref{tb2} illustrates the performance of our $\text{AI-CNet3D}_H$ on both widefield (Dataset 1) and ONH-only (Dataset 2) volumes. The 3D CNN proposed by \cite{Maetschke_2019} was optimized for Dataset 2 volumes. However, its performance degrades significantly when evaluated on Dataset 1, indicating limited generalizability across datasets. Our improvement in performance indicates that the advantage of our approach is achieved by embedding our understanding of disease pathology through modeling and thus is not limited to particular scans and orientations. Our approach removes the spatial constraints that may limit models' ability to analyze 3D images and allows for computation of biologically-meaningful cross-correlations between regions that are tailored toward a specific clinical question. This is a key advantage, as it offers a customization approach that can be applied to multiple imaging modalities and disease settings.

\subsection{Multi-task Fine-tuning to Enforce Visualization-Based Consistency}
\label{mtft}

Qualitative results in Fig. \ref{fig:heatmap} demonstrate strong alignment and consistency between the last attention and convolutional layers, suggesting effective information sharing between them. This interaction enables the model to leverage the strengths of both mechanisms: convolutions excel at capturing fine-grained local structures due to their limited receptive fields, while attention mechanisms provide a complementary ability to model long-range dependencies. By aligning their outputs, the convolutional layer gains access to broader contextual information, while the attention mechanism benefits from enhanced local feature sensitivity. This synergy improves feature representation, allowing the model to integrate both fine-detail spatial structures and global contextual cues, ultimately enhancing interpretability and performance.

Fine-tuning the model with a combination of unsupervised visualization consistency and BCE losses plays a crucial role in enhancing its generalization ability. By enforcing consistency in model visualizations across attention and convolutional layers, the model learns more stable and robust feature representations, reducing susceptibility to spurious correlations in the data. The inclusion of BCE loss during fine-tuning ensures that the model retains its learned features for classification while preventing degeneration. We can also notice that regardless of the attention mechanism used in Table \ref{tb3}, multi-task fine-tuning improves individual results. This shows how our fine-tuning strategy can be used for general improvement after a model has been trained. This improvement is particularly significant for medical imaging tasks: by leveraging consistency-based multi-task fine-tuning, we offer a dependable AI-assisted analysis framework that is agnostic to attention model backbone, enabling consistent performance improvement and interpretability enhancement, critical for clinical adoption.

\subsection{Computational Complexity}
\label{compcomplx}
Model efficiency is a crucial consideration in developing 3D models, particularly for applications in low-resource settings where hardware limitations and inference speed are major constraints, as noted in prior work such as \cite{unetrpp}. In designing our 3D cross-attention mechanism, we prioritized a lightweight architecture that balances robust performance with computational efficiency, allowing for fast deployment and real-time processing on resource-limited devices.

\begin{figure}[htb]

\begin{minipage}[b]{1.0\linewidth}
  \centering
  \centerline{\includegraphics[width=\textwidth]{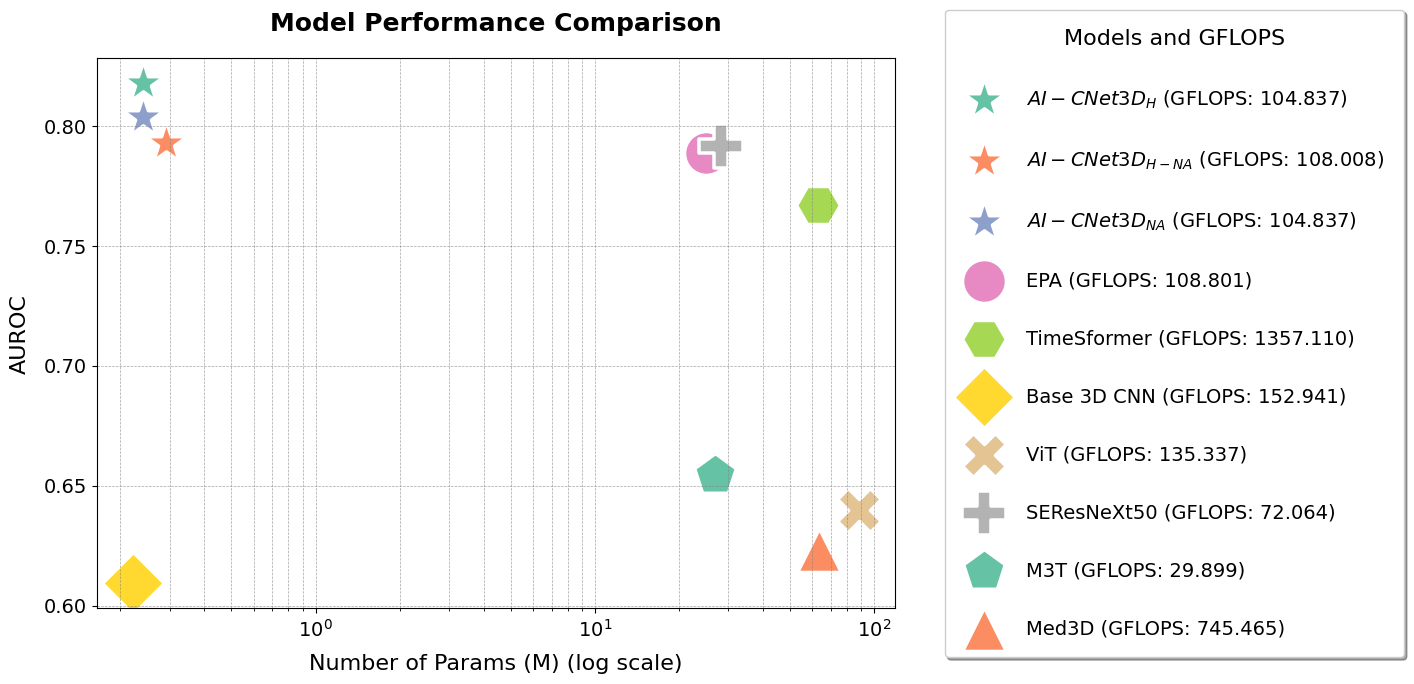}}
\end{minipage}
\caption{Efficiency analysis of our models and baseline models with parameters displayed on a log scale for enhanced visualization. Our models achieve the highest AUROC, matching EPA’s GFLOP performance while significantly reducing parameter count, highlighting their efficiency and accuracy.}
\label{fig:complexity}
\end{figure}

Our 3D cross-attention mechanism achieves these efficiency gains primarily by eliminating the need for volume projections for spatial attention computation, which drastically reduces parameter count without compromising model accuracy. As shown in Figure \ref{fig:complexity}, when evaluated on 128×192×112 pixel volumes with a single-batch input, traditional models exhibit significant variation in computational demands. The Base 3D CNN operates with approximately 222k parameters and 152.941 GFLOPS, while SEResNeXt50, EPA, and M3T require 28.36M, 25.01M, and 27.05M parameters, with computational costs of 72.064, 108.801, and 29.90 GFLOPS, respectively. More resource-intensive models like TimeSformer, ViT, and Med3D demand 63.15M, 88.18M, 63.30M parameters, with significantly higher computational costs of 1357.11, 135.34, 745.47 GFLOPS, respectively. In contrast, our proposed $AI\text{-}CNet3D_H$, $AI\text{-}CNet3D_{H-NA}$, and $AI\text{-}CNet3D_{NA}$ achieve competitive performance with just 241k to 291k parameters while maintaining computational efficiency at 104.837 to 108.008 GFLOPS, underscoring our model's optimized architecture and computational efficiency.

Our method's ability to outperform models like TimeSformer in terms of computational efficiency position it as a highly-scalable approach, especially as medical imaging datasets grow larger and more complex. The reduced resource demand also makes AI-CNet3D ideal for integration into portable medical devices and remote healthcare, offering real-time, reliable diagnostics in resource-constrained settings. 


\subsection{3D Visualization}
\label{vizdisc}

\begin{figure*}[htb]

\begin{minipage}[b]{1.0\linewidth}
  \centering
  \centerline{\includegraphics[width=\textwidth]{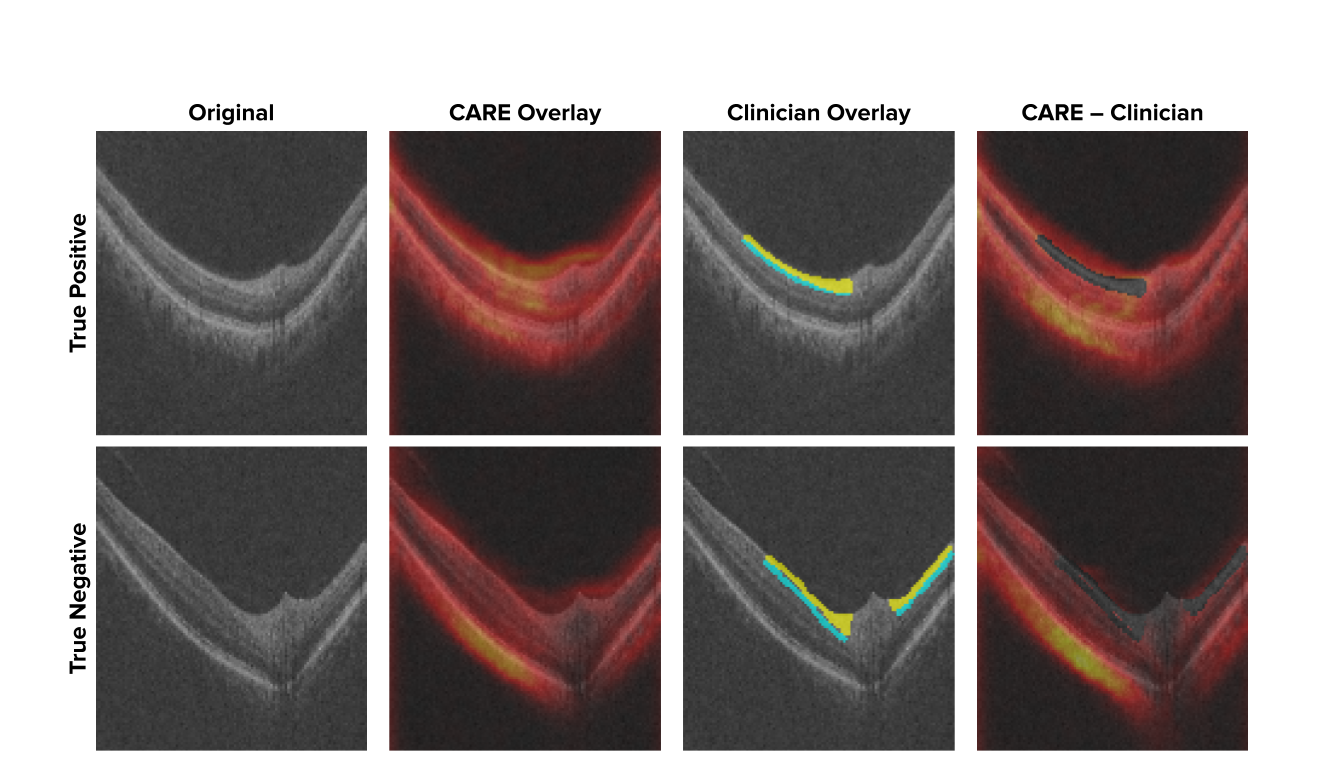}}
\end{minipage}
\caption{A comparison of what our CARE highlights versus the important regions identified by the clinician. In the final column, we subtract the clinical regions of interest (RoIs) from the CARE heatmap to highlight how our method captures additional features within the volumes beyond the clinical ROIs.}
\label{fig:clinician_overlay}
\end{figure*}

CARE visualizations suggest that leveraging 3D OCT volumes for glaucoma diagnosis enables deep learning models incorporating attention to learn information from deeper features in the retina beyond the typical RNFL or GCC relied on clinically in 2D OCT reports. While we have leveraged this advantage in the X-Z plane to correlate nerve to macula or superior to inferior as discussed above, such advantage may equally apply in the Y direction. 

Past work has shown that Grad-CAM successfully highlights the RNFL layer \citep{thakoor2020robust}, consistent with our current understanding of glaucoma pathophysiology. Both Grad-CAM and CARE demonstrated the importance of RNFL and GCC layers, as showcased by involvement of these superficial layers in the model's decision making. However, only CARE demonstrated the use of features from deeper retinal layers for classification of glaucoma, which deviates from the traditional understanding of glaucoma. Previous literature has suggested possible anatomical changes in the photoreceptor layers in patients with glaucoma \citep{fan2011measurement, trolli2024parafoveal}, although this has not been validated or used in clinical settings. Our results are consistent with these studies, indicating that deeper retinal structures, such as the photoreceptor layer, may have subtle changes that can be leveraged by AI models to detect glaucoma.

The power of multi-task fine-tuning (MTFT) in allowing convolutional layers (as visualized by Grad-CAM) and attention layers (as visualized by CARE) to share key information is also well demonstrated by our visualization. The consistency between the visualizations of the last convolutional and attention layers illustrates the successful integration of the shared information, as discussed in \ref{mtft}. The interpretability of this visualization is further supported by increased signals originating from the superior and inferior rims of the optic nerve head, following an arcuate pattern into the macula (especially evident in Topcon true-negative coronal slices in Fig. \ref{fig:heatmap}), consistent with our clinical understanding of glaucoma.

\begin{table}[h]
\centering
\begin{tabular}{lc}
\hline
\textbf{Metric Avg. ± Std.} & \textbf{Value} \\
\hline
Mask Coverage & 0.9337 ± 0.1004 \\
CARE Coverage & 0.0293 ± 0.0125\\
Enrichment & 3.0915 ± 0.6144\\
\hline
\end{tabular}
\caption{Quantitative evaluation of CARE attention alignment with clinician-annotated anatomical regions. Mask coverage measures the fraction of clinician-defined regions that receive high CARE attention, CARE coverage indicates the fraction of high-attention pixels that fall within anatomical boundaries, and enrichment quantifies how much more concentrated CARE is within versus outside the expert-annotated RNFL/GCL regions of interest.}
\label{tab:attention_metrics}
\end{table}

To validate our method's ability to capture clinically relevant anatomical regions, we obtained expert segmentation annotations from a clinician at Columbia University Irving Medical Center. The clinician manually segmented the middle slice of each test volume from the Topcon dataset, identifying RNFL and GCL Regions of Interest (RoIs). These annotations were converted into binary masks and compared against the corresponding middle slice of our CARE attention maps generated using the first attention layer of $AI\text{-}CNet3D_H$ after multi-task fine-tuning. We converted each CARE slice into a binary segmentation using the 75th percentile as a threshold ($\tau$) to identify the most intense regions. We then computed mask coverage as $\frac{|M \cap \text{CARE}_{\tau}|}{|M|}$, CARE coverage as $\frac{|M \cap \text{CARE}_{\tau}|}{|\text{CARE}_{\tau}|}$, and enrichment as $\frac{\bar{\text{CARE}}_M}{\bar{\text{CARE}}_{V \setminus M}}$, where $M$ represents the clinician-annotated mask, $\text{CARE}_{\tau}$ represents the binarized CARE map above threshold $\tau$, $M \cap \text{CARE}_{\tau}$ represents their intersection, $V$ represents the full slice, and $\bar{\text{CARE}}_M$ and $\bar{\text{CARE}}_{V \setminus M}$ denote the mean CARE intensity inside and outside the annotation mask area, respectively. Values of 1.0 for enrichment indicate the same CARE density inside and outside the mask, while values greater than 1.0 indicate that CARE is more enriched inside the annotation mask area.

Table \ref{tab:attention_metrics} reveals that our model demonstrates exceptional annotation mask coverage in targeting anatomical structures, with 93.37\% of anatomical pixels receiving significant attention. As seen with CARE coverage, only 2.93\% of the highest attention pixels fall within the clinician-defined anatomical boundaries. This reflects our model's broader analytical scope, with potential to enable novel biomarker discovery by capturing extensive retinal features beyond conventionally-defined clinical RoIs. Additionally, the CARE coverage is likely slightly higher in reality, as the physician annotation of RoIs were limited by resolution, and the true RoIs (RNFL and GCL layers) are anatomically known to continue beyond the annotated regions, nasally and temporally. At the same time, the fact that the enrichment value for all test cases is greater than 1.0, indicates that 100\% of the time, CARE is more concentrated within the clinically defined regions of interest. This comprehensive attention distribution is visualized in Figure \ref{fig:clinician_overlay}, which demonstrates that our model encompasses the clinician's primary areas of interest while extending analysis to capture additional diagnostically-relevant features throughout the retinal volume.

There are some caveats to using CARE and Grad-CAM for consistency and 3D visualization. Notably, as shown in Fig. \ref{fig:heatmap}, consistency after multi-task fine-tuning (MTFT) is significantly stronger in the last layers for true negative cases. This may explain the larger increases in specificity observed in Table \ref{tb3}, as the model becomes more adept at learning from negative cases. Additionally, a qualitative comparison of Zeiss and Topcon examples reveals that Zeiss visualizations are less precise. This is likely due to the lower resolution of Zeiss volumes (64×128×64), which leads to further degradation in heatmap quality when downsampled within our model.


\section{Conclusions and Future Directions}

In this paper, we introduced a novel 3D cross-attention mechanism for widefield OCT reports, demonstrating its potential to enhance glaucoma classification by incorporating anatomical priors and inter-region correlations within 3D volumes. Our approach, implemented as the $\text{AI-CNet3D}$ model, outperforms traditional 3D CNN-based models, as well as other state-of-the-art spatial and channel attention methods, in both classification performance and computational efficiency. The use of superior-inferior and macula-nerve cross-attention mechanisms allows our model to leverage the inherent anatomical relationships in OCT scans, which leads to improved sensitivity, specificity, and overall performance, which may be beneficial in the detection of subtle or early-stage glaucoma. The computational efficiency of our model, achieved by eliminating the need for volume projections and reducing parameter count, makes it particularly suitable for resource-constrained environments, such as portable imaging devices and point-of-care diagnostics.

Our newly-introduced CARE technique enables visualization of 3D attention mechanisms, which are inherently challenging to portray visually. Compared to Grad-CAM, CARE visualizations highlighted the RNFL and GCC as well as deeper structures, such as photoreceptors, as critical for the model's decision-making. We enhanced model performance by training with a joint unsupervised visualization consistency loss and regular BCE loss as a fine-tuning step. This process allows the model to generalize better while also increasing its interpretability with alignment between layers. Furthermore, CARE specifically elucidated the contribution from deeper retinal structures for glaucoma classification. By leveraging our technique, both CARE and 3D Grad-CAM can be applied to hybrid CNN-attention models, empowering clinicians to interpret contributions of every layer in the model, including attention layers, rather than limiting their insights to convolutional layers alone.


Our work also contributes to the growing body of research on 3D deep learning models for medical imaging, highlighting the importance of attention-based and consistency-based multi-task fine-tuning approaches to improve interpretability and model performance in clinically relevant tasks. To ensure class balance, we used equal amounts of glaucomatous and non-glaucomatous data in our training pipeline, both with and without fine-tuning. As a direction for future work, the full set of available samples from both datasets, or additional data from external domains, could be incorporated into a pre-training task to improve model robustness and generalization. Additionally, we aim to further investigate the clinical impact of our model by leveraging CARE outputs to create pseudo-segmentation masks that can be corrected by minimal annotation of a clinician. We can then use our cross-attention network to identify and segment novel biomarkers for retinal diseases, thereby advancing AI-driven diagnostics in ophthalmology.


\acks{The authors are grateful to George A. Cioffi, Jeffrey M. Liebmann, Aakriti G. Shukla, and Ives A. Valenzuela for their guidance and insights as glaucoma specialists. Many thanks go to Mary Durbin and Reena Chopra (Topcon Healthcare, Inc.) for collaborative data sharing as well as Emmanouil Tsamis and Donald C. Hood for providing ground truth labels for the volumetric OCT data. We also acknowledge our funding sources: Columbia University's Data Science Institute Seed Fund and an Unrestricted Grant from Research to Prevent Blindness, New York, NY, USA.}

%
\ethics{Our Topcon Dataset 1 was from a retrospective study conducted in accordance with the Declaration of Helsinki and approved on 14 February 2023 by Advarra Institutional Review Board (MOD01564217).}

\coi{Author K.A.T. has received research funding from Topcon Healthcare for a study unrelated to the topic of this paper. All other authors declare no further conflicts of interest.}

\data{Given the sensitive nature of the human subject data used in our study, it is essential to maintain confidentiality and adhere to ethical guidelines. While our data collection received approval from the Institutional Review Board (IRB), any further access to Dataset 1 must be carefully regulated. To ensure appropriate use, we would need to assess data requests to verify that the intended purpose aligns with the submitted request. For the journal’s internal evaluation, we have already provided a minimal de-identified dataset through article figures and can furnish additional datasets if further review is necessary. Dataset 2 is available publicly as published by \cite{Maetschke_2019} at \href{https://zenodo.org/records/1481223}{https://zenodo.org/records/1481223}.}

\bibliography{sample}

\begin{thebibliography}{56}
\providecommand{\natexlab}[1]{#1}
\providecommand{\url}[1]{\texttt{#1}}
\expandafter\ifx\csname urlstyle\endcsname\relax
  \providecommand{\doi}[1]{doi: #1}\else
  \providecommand{\doi}{doi: \begingroup \urlstyle{rm}\Url}\fi

\bibitem[Abnar and Zuidema(2020)]{abnar2020quantifying}
Samira Abnar and Willem Zuidema.
\newblock Quantifying attention flow in transformers.
\newblock \emph{arXiv preprint arXiv:2005.00928}, 2020.

\bibitem[Avram et~al.(2023)Avram, Durmus, Rakocz, Corradetti, An, Nitalla, Rudas, Wakatsuki, Hirabayashi, Velaga, et~al.]{avram2023slivit}
Oren Avram, Berkin Durmus, Nadav Rakocz, Giulia Corradetti, Ulzee An, Muneeswar~G Nitalla, {\'A}kos Rudas, Yu~Wakatsuki, Kazutaka Hirabayashi, Swetha Velaga, et~al.
\newblock Slivit: a general ai framework for clinical-feature diagnosis from limited 3d biomedical-imaging data.
\newblock \emph{Research Square}, pages rs--3, 2023.

\bibitem[Barros et~al.(2020)Barros, Moura, Freire, Taleb, Valentim, and Morais]{barros2020machine}
Daniele~MS Barros, Julio~CC Moura, Cefas~R Freire, Alexandre~C Taleb, Ricardo~AM Valentim, and Philippi~SG Morais.
\newblock Machine learning applied to retinal image processing for glaucoma detection: review and perspective.
\newblock \emph{Biomedical engineering online}, 19:\penalty0 1--21, 2020.

\bibitem[Bertasius et~al.(2021)Bertasius, Wang, and Torresani]{bertasius2021space}
Gedas Bertasius, Heng Wang, and Lorenzo Torresani.
\newblock Is space-time attention all you need for video understanding?
\newblock In \emph{ICML}, volume~2, page~4, 2021.

\bibitem[Bock et~al.(2010)Bock, Meier, Ny{\'u}l, Hornegger, and Michelson]{bock2010glaucoma}
R{\"u}diger Bock, J{\"o}rg Meier, L{\'a}szl{\'o}~G Ny{\'u}l, Joachim Hornegger, and Georg Michelson.
\newblock Glaucoma risk index: automated glaucoma detection from color fundus images.
\newblock \emph{Medical image analysis}, 14\penalty0 (3):\penalty0 471--481, 2010.

\bibitem[Bussel et~al.(2014)Bussel, Wollstein, and Schuman]{bussel2014oct}
Igor~I Bussel, Gadi Wollstein, and Joel~S Schuman.
\newblock Oct for glaucoma diagnosis, screening and detection of glaucoma progression.
\newblock \emph{British Journal of Ophthalmology}, 98\penalty0 (Suppl 2):\penalty0 ii15--ii19, 2014.

\bibitem[Cao et~al.(2022)Cao, Wang, Chen, Jiang, Zhang, Tian, and Wang]{cao2022swin}
Hu~Cao, Yueyue Wang, Joy Chen, Dongsheng Jiang, Xiaopeng Zhang, Qi~Tian, and Manning Wang.
\newblock Swin-unet: Unet-like pure transformer for medical image segmentation.
\newblock In \emph{European conference on computer vision}, pages 205--218. Springer, 2022.

\bibitem[Chefer et~al.(2021)Chefer, Gur, and Wolf]{chefer2021transformer}
Hila Chefer, Shir Gur, and Lior Wolf.
\newblock Transformer interpretability beyond attention visualization.
\newblock In \emph{Proceedings of the IEEE/CVF conference on computer vision and pattern recognition}, pages 782--791, 2021.

\bibitem[Chen et~al.(2021)Chen, Fan, and Panda]{chen2021crossvit}
Chun-Fu~Richard Chen, Quanfu Fan, and Rameswar Panda.
\newblock Crossvit: Cross-attention multi-scale vision transformer for image classification.
\newblock In \emph{Proceedings of the IEEE/CVF international conference on computer vision}, pages 357--366, 2021.

\bibitem[Chen et~al.(2019)Chen, Ma, and Zheng]{chen2019med3d}
Sihong Chen, Kai Ma, and Yefeng Zheng.
\newblock Med3d: Transfer learning for 3d medical image analysis.
\newblock \emph{arXiv preprint arXiv:1904.00625}, 2019.

\bibitem[Chen et~al.(2018)Chen, Hoguet, Junk, Nouri-Mahdavi, Radhakrishnan, Takusagawa, and Chen]{chen2018spectral}
Teresa~C Chen, Ambika Hoguet, Anna~K Junk, Kouros Nouri-Mahdavi, Sunita Radhakrishnan, Hana~L Takusagawa, and Philip~P Chen.
\newblock Spectral-domain oct: helping the clinician diagnose glaucoma: a report by the american academy of ophthalmology.
\newblock \emph{Ophthalmology}, 125\penalty0 (11):\penalty0 1817--1827, 2018.

\bibitem[Chen et~al.(2020)Chen, Kornblith, Norouzi, and Hinton]{chen2020simple}
Ting Chen, Simon Kornblith, Mohammad Norouzi, and Geoffrey Hinton.
\newblock A simple framework for contrastive learning of visual representations.
\newblock In \emph{International conference on machine learning}, pages 1597--1607. PMLR, 2020.

\bibitem[Chen et~al.(2015)Chen, Xu, Kee~Wong, Wong, and Liu]{7318462}
Xiangyu Chen, Yanwu Xu, Damon~Wing Kee~Wong, Tien~Yin Wong, and Jiang Liu.
\newblock Glaucoma detection based on deep convolutional neural network.
\newblock In \emph{2015 37th Annual International Conference of the IEEE Engineering in Medicine and Biology Society (EMBC)}, pages 715--718, 2015.
\newblock \doi{10.1109/EMBC.2015.7318462}.

\bibitem[Dosovitskiy et~al.(2020)Dosovitskiy, Beyer, Kolesnikov, Weissenborn, Zhai, Unterthiner, Dehghani, Minderer, Heigold, Gelly, Uszkoreit, and Houlsby]{vit}
Alexey Dosovitskiy, Lucas Beyer, Alexander Kolesnikov, Dirk Weissenborn, Xiaohua Zhai, Thomas Unterthiner, Mostafa Dehghani, Matthias Minderer, Georg Heigold, Sylvain Gelly, Jakob Uszkoreit, and Neil Houlsby.
\newblock An image is worth 16x16 words: Transformers for image recognition at scale.
\newblock \emph{CoRR}, abs/2010.11929, 2020.
\newblock URL \url{https://arxiv.org/abs/2010.11929}.

\bibitem[Fan et~al.(2011)Fan, Huang, Lam, and Leung]{fan2011measurement}
Ning Fan, Nina Huang, Dennis Shun~Chiu Lam, and Christopher Kai-shun Leung.
\newblock Measurement of photoreceptor layer in glaucoma: a spectral-domain optical coherence tomography study.
\newblock \emph{Journal of ophthalmology}, 2011\penalty0 (1):\penalty0 264803, 2011.

\bibitem[Geevarghese et~al.(2021)Geevarghese, Wollstein, Ishikawa, and Schuman]{geevarghese2021optical}
Alexi Geevarghese, Gadi Wollstein, Hiroshi Ishikawa, and Joel~S Schuman.
\newblock Optical coherence tomography and glaucoma.
\newblock \emph{Annual review of vision science}, 7\penalty0 (1):\penalty0 693--726, 2021.

\bibitem[George et~al.(2020)George, Antony, Ishikawa, Wollstein, Schuman, and Garnavi]{9112644}
Yasmeen George, Bhavna~J. Antony, Hiroshi Ishikawa, Gadi Wollstein, Joel~S. Schuman, and Rahil Garnavi.
\newblock Attention-guided 3d-cnn framework for glaucoma detection and structural-functional association using volumetric images.
\newblock \emph{IEEE Journal of Biomedical and Health Informatics}, 24\penalty0 (12):\penalty0 3421--3430, 2020.
\newblock \doi{10.1109/JBHI.2020.3001019}.

\bibitem[Grill et~al.(2020)Grill, Strub, Altch{\'e}, Tallec, Richemond, Buchatskaya, Doersch, Avila~Pires, Guo, Gheshlaghi~Azar, et~al.]{grill2020bootstrap}
Jean-Bastien Grill, Florian Strub, Florent Altch{\'e}, Corentin Tallec, Pierre Richemond, Elena Buchatskaya, Carl Doersch, Bernardo Avila~Pires, Zhaohan Guo, Mohammad Gheshlaghi~Azar, et~al.
\newblock Bootstrap your own latent-a new approach to self-supervised learning.
\newblock \emph{Advances in neural information processing systems}, 33:\penalty0 21271--21284, 2020.

\bibitem[Gu and Dao(2023)]{gu2023mamba}
Albert Gu and Tri Dao.
\newblock Mamba: Linear-time sequence modeling with selective state spaces.
\newblock \emph{arXiv preprint arXiv:2312.00752}, 2023.

\bibitem[Gu et~al.(2021)Gu, Goel, and R{\'e}]{gu2021efficiently}
Albert Gu, Karan Goel, and Christopher R{\'e}.
\newblock Efficiently modeling long sequences with structured state spaces.
\newblock \emph{arXiv preprint arXiv:2111.00396}, 2021.

\bibitem[Guo et~al.(2019)Guo, Zheng, Fan, Yu, and Wang]{guo2019visual}
Hao Guo, Kang Zheng, Xiaochuan Fan, Hongkai Yu, and Song Wang.
\newblock Visual attention consistency under image transforms for multi-label image classification.
\newblock In \emph{Proceedings of the IEEE/CVF conference on computer vision and pattern recognition}, pages 729--739, 2019.

\bibitem[Hood et~al.(2022)Hood, La~Bruna, Tsamis, Thakoor, Rai, Leshno, de~Moraes, Cioffi, and Liebmann]{hood2022detecting}
Donald~C Hood, Sol La~Bruna, Emmanouil Tsamis, Kaveri~A Thakoor, Anvit Rai, Ari Leshno, Carlos~GV de~Moraes, George~A Cioffi, and Jeffrey~M Liebmann.
\newblock Detecting glaucoma with only oct: Implications for the clinic, research, screening, and ai development.
\newblock \emph{Progress in Retinal and Eye Research}, 90:\penalty0 101052, 2022.

\bibitem[Hu et~al.(2018)Hu, Shen, and Sun]{squeezeexcite}
Jie Hu, Li~Shen, and Gang Sun.
\newblock Squeeze-and-excitation networks.
\newblock In \emph{2018 IEEE/CVF Conference on Computer Vision and Pattern Recognition}, pages 7132--7141, 2018.
\newblock \doi{10.1109/CVPR.2018.00745}.

\bibitem[Islam et~al.(2020)Islam, Vibashan, Jose, Wijethilake, Utkarsh, and Ren]{islam2020brain}
Mobarakol Islam, VS~Vibashan, V~Jeya~Maria Jose, Navodini Wijethilake, Uppal Utkarsh, and Hongliang Ren.
\newblock Brain tumor segmentation and survival prediction using 3d attention unet.
\newblock In \emph{Brainlesion: Glioma, Multiple Sclerosis, Stroke and Traumatic Brain Injuries: 5th International Workshop, BrainLes 2019, Held in Conjunction with MICCAI 2019, Shenzhen, China, October 17, 2019, Revised Selected Papers, Part I 5}, pages 262--272. Springer, 2020.

\bibitem[Jang and Hwang(2022)]{jang2022m3t}
Jinseong Jang and Dosik Hwang.
\newblock M3t: three-dimensional medical image classifier using multi-plane and multi-slice transformer.
\newblock In \emph{Proceedings of the IEEE/CVF conference on computer vision and pattern recognition}, pages 20718--20729, 2022.

\bibitem[Li et~al.(2019)Li, Xu, Wang, Jiang, and Liu]{li2019attention}
Liu Li, Mai Xu, Xiaofei Wang, Lai Jiang, and Hanruo Liu.
\newblock Attention based glaucoma detection: A large-scale database and cnn model.
\newblock In \emph{Proceedings of the IEEE/CVF conference on computer vision and pattern recognition}, pages 10571--10580, 2019.

\bibitem[Li et~al.(2020)Li, Kan, and He]{li2020unsupervised}
Yang Li, Shichao Kan, and Zhihai He.
\newblock Unsupervised deep metric learning with transformed attention consistency and contrastive clustering loss.
\newblock In \emph{European Conference on Computer Vision}, pages 141--157. Springer, 2020.

\bibitem[Lin et~al.(2022)Lin, Cheng, Wu, and Shen]{lin2022cat}
Hezheng Lin, Xing Cheng, Xiangyu Wu, and Dong Shen.
\newblock Cat: Cross attention in vision transformer.
\newblock In \emph{2022 IEEE international conference on multimedia and expo (ICME)}, pages 1--6. IEEE, 2022.

\bibitem[Liu et~al.(2024{\natexlab{a}})Liu, Yang, Zhou, Yu, Liang, Yu, Zhang, Zheng, and Wang]{liu2024swin}
Jiarun Liu, Hao Yang, Hong-Yu Zhou, Lequan Yu, Yong Liang, Yizhou Yu, Shaoting Zhang, Hairong Zheng, and Shanshan Wang.
\newblock Swin-umamba†: Adapting mamba-based vision foundation models for medical image segmentation.
\newblock \emph{IEEE Transactions on Medical Imaging}, 2024{\natexlab{a}}.

\bibitem[Liu et~al.(2024{\natexlab{b}})Liu, Xu, Woicik, Shapiro, Blazes, Wu, Steffen, Cukras, Lee, Zhang, Lee, and Wang]{liu2024octcubem3dmultimodaloptical}
Zixuan Liu, Hanwen Xu, Addie Woicik, Linda~G. Shapiro, Marian Blazes, Yue Wu, Verena Steffen, Catherine Cukras, Cecilia~S. Lee, Miao Zhang, Aaron~Y. Lee, and Sheng Wang.
\newblock Octcube-m: A 3d multimodal optical coherence tomography foundation model for retinal and systemic diseases with cross-cohort and cross-device validation, 2024{\natexlab{b}}.
\newblock URL \url{https://arxiv.org/abs/2408.11227}.

\bibitem[Luo et~al.(2023)Luo, Shi, Tian, Elze, and Wang]{luo2023harvard}
Yan Luo, Min Shi, Yu~Tian, Tobias Elze, and Mengyu Wang.
\newblock Harvard glaucoma detection and progression: A multimodal multitask dataset and generalization-reinforced semi-supervised learning.
\newblock In \emph{Proceedings of the IEEE/CVF International Conference on Computer Vision}, pages 20471--20482, 2023.

\bibitem[Maetschke et~al.(2019)Maetschke, Antony, Ishikawa, Wollstein, Schuman, and Garnavi]{Maetschke_2019}
Stefan Maetschke, Bhavna Antony, Hiroshi Ishikawa, Gadi Wollstein, Joel Schuman, and Rahil Garnavi.
\newblock A feature agnostic approach for glaucoma detection in oct volumes.
\newblock \emph{PLOS ONE}, 14\penalty0 (7):\penalty0 e0219126, July 2019.
\newblock ISSN 1932-6203.
\newblock \doi{10.1371/journal.pone.0219126}.
\newblock URL \url{http://dx.doi.org/10.1371/journal.pone.0219126}.

\bibitem[McMahan et~al.(2017)McMahan, Moore, Ramage, Hampson, and y~Arcas]{mcmahan2017communication}
Brendan McMahan, Eider Moore, Daniel Ramage, Seth Hampson, and Blaise~Aguera y~Arcas.
\newblock Communication-efficient learning of deep networks from decentralized data.
\newblock In \emph{Artificial intelligence and statistics}, pages 1273--1282. PMLR, 2017.

\bibitem[Mehta et~al.(2021)Mehta, Petersen, Wen, Banitt, Chen, Bojikian, Egan, Lee, Balazinska, Lee, et~al.]{mehta2021automated}
Parmita Mehta, Christine~A Petersen, Joanne~C Wen, Michael~R Banitt, Philip~P Chen, Karine~D Bojikian, Catherine Egan, Su-In Lee, Magdalena Balazinska, Aaron~Y Lee, et~al.
\newblock Automated detection of glaucoma with interpretable machine learning using clinical data and multimodal retinal images.
\newblock \emph{American Journal of Ophthalmology}, 231:\penalty0 154--169, 2021.

\bibitem[Mirzazadeh et~al.(2023)Mirzazadeh, Dubost, Pike, Maniar, Zuo, Lee-Messer, and Rubin]{mirzazadeh2023atcon}
Ali Mirzazadeh, Florian Dubost, Maxwell Pike, Krish Maniar, Max Zuo, Christopher Lee-Messer, and Daniel Rubin.
\newblock Atcon: Attention consistency for vision models.
\newblock In \emph{Proceedings of the IEEE/CVF Winter Conference on Applications of Computer Vision}, pages 1880--1889, 2023.

\bibitem[Pang et~al.(2023)Pang, Liang, Huang, Chen, Li, Li, Huang, and Wang]{pang2023slim}
Yan Pang, Jiaming Liang, Teng Huang, Hao Chen, Yunhao Li, Dan Li, Lin Huang, and Qiong Wang.
\newblock Slim unetr: scale hybrid transformers to efficient 3d medical image segmentation under limited computational resources.
\newblock \emph{IEEE Transactions on Medical Imaging}, 43\penalty0 (3):\penalty0 994--1005, 2023.

\bibitem[Quigley and Broman(2006)]{quigley2006number}
Harry~A Quigley and Aimee~T Broman.
\newblock The number of people with glaucoma worldwide in 2010 and 2020.
\newblock \emph{British journal of ophthalmology}, 90\penalty0 (3):\penalty0 262--267, 2006.

\bibitem[Selvaraju et~al.(2020)Selvaraju, Cogswell, Das, Vedantam, Parikh, and Batra]{selvaraju2020grad}
Ramprasaath~R Selvaraju, Michael Cogswell, Abhishek Das, Ramakrishna Vedantam, Devi Parikh, and Dhruv Batra.
\newblock Grad-cam: visual explanations from deep networks via gradient-based localization.
\newblock \emph{International journal of computer vision}, 128:\penalty0 336--359, 2020.

\bibitem[Shaikhina and Khovanova(2017)]{shaikhina2017handling}
Torgyn Shaikhina and Natalia~A Khovanova.
\newblock Handling limited datasets with neural networks in medical applications: A small-data approach.
\newblock \emph{Artificial intelligence in medicine}, 75:\penalty0 51--63, 2017.

\bibitem[Shaker et~al.(2023)Shaker, Maaz, Rasheed, Khan, Yang, and Khan]{shaker2023swiftformer}
Abdelrahman Shaker, Muhammad Maaz, Hanoona Rasheed, Salman Khan, Ming-Hsuan Yang, and Fahad~Shahbaz Khan.
\newblock Swiftformer: Efficient additive attention for transformer-based real-time mobile vision applications.
\newblock In \emph{Proceedings of the IEEE/CVF international conference on computer vision}, pages 17425--17436, 2023.

\bibitem[Shaker et~al.(2024)Shaker, Maaz, Rasheed, Khan, Yang, and Khan]{unetrpp}
Abdelrahman~M. Shaker, Muhammad Maaz, Hanoona Rasheed, Salman Khan, Ming-Hsuan Yang, and Fahad~Shahbaz Khan.
\newblock Unetr++: Delving into efficient and accurate 3d medical image segmentation.
\newblock \emph{IEEE Transactions on Medical Imaging}, 2024.
\newblock \doi{10.1109/TMI.2024.3398728}.

\bibitem[Steinmetz et~al.(2021)Steinmetz, Bourne, Briant, Flaxman, Taylor, Jonas, Abdoli, Abrha, Abualhasan, Abu-Gharbieh, et~al.]{steinmetz2021causes}
Jaimie~D Steinmetz, Rupert~RA Bourne, Paul~Svitil Briant, Seth~R Flaxman, Hugh~RB Taylor, Jost~B Jonas, Amir~Aberhe Abdoli, Woldu~Aberhe Abrha, Ahmed Abualhasan, Eman~Girum Abu-Gharbieh, et~al.
\newblock Causes of blindness and vision impairment in 2020 and trends over 30 years, and prevalence of avoidable blindness in relation to vision 2020: the right to sight: an analysis for the global burden of disease study.
\newblock \emph{The Lancet Global Health}, 9\penalty0 (2):\penalty0 e144--e160, 2021.

\bibitem[Tang et~al.(2022)Tang, Yang, Li, Roth, Landman, Xu, Nath, and Hatamizadeh]{tang2022self}
Yucheng Tang, Dong Yang, Wenqi Li, Holger~R Roth, Bennett Landman, Daguang Xu, Vishwesh Nath, and Ali Hatamizadeh.
\newblock Self-supervised pre-training of swin transformers for 3d medical image analysis.
\newblock In \emph{Proceedings of the IEEE/CVF conference on computer vision and pattern recognition}, pages 20730--20740, 2022.

\bibitem[Thakoor et~al.(2020)Thakoor, Koorathota, Hood, and Sajda]{thakoor2020robust}
Kaveri~A Thakoor, Sharath~C Koorathota, Donald~C Hood, and Paul Sajda.
\newblock Robust and interpretable convolutional neural networks to detect glaucoma in optical coherence tomography images.
\newblock \emph{IEEE Transactions on Biomedical Engineering}, 68\penalty0 (8):\penalty0 2456--2466, 2020.

\bibitem[Tran et~al.(2015)Tran, Bourdev, Fergus, Torresani, and Paluri]{tran2015learning}
Du~Tran, Lubomir Bourdev, Rob Fergus, Lorenzo Torresani, and Manohar Paluri.
\newblock Learning spatiotemporal features with 3d convolutional networks.
\newblock In \emph{Proceedings of the IEEE international conference on computer vision}, pages 4489--4497, 2015.

\bibitem[Trolli et~al.(2024)Trolli, Roda, Valsecchi, Cacciatore, Nardi, Della~Pasqua, Mercanti, and Fontana]{trolli2024parafoveal}
Eleonora Trolli, Matilde Roda, Nicola Valsecchi, Davide Cacciatore, Elena Nardi, Valentina Della~Pasqua, Andrea Mercanti, and Luigi Fontana.
\newblock A parafoveal retinal cones analysis using adaptive-optics retinal camera in patients with primary open angle glaucoma.
\newblock \emph{Eye}, pages 1--7, 2024.

\bibitem[Vaswani(2017)]{vaswani2017attention}
A~Vaswani.
\newblock Attention is all you need.
\newblock \emph{Advances in Neural Information Processing Systems}, 2017.

\bibitem[Wang et~al.(2019{\natexlab{a}})Wang, Wu, Karanam, Peng, Singh, Liu, and Metaxas]{wang2019sharpen}
Lezi Wang, Ziyan Wu, Srikrishna Karanam, Kuan-Chuan Peng, Rajat~Vikram Singh, Bo~Liu, and Dimitris~N Metaxas.
\newblock Sharpen focus: Learning with attention separability and consistency.
\newblock In \emph{Proceedings of the IEEE/CVF International Conference on Computer Vision}, pages 512--521, 2019{\natexlab{a}}.

\bibitem[Wang et~al.(2019{\natexlab{b}})Wang, Han, Chen, Gao, and Vasconcelos]{wang2019volumetric}
Xudong Wang, Shizhong Han, Yunqiang Chen, Dashan Gao, and Nuno Vasconcelos.
\newblock Volumetric attention for 3d medical image segmentation and detection.
\newblock In \emph{Medical Image Computing and Computer Assisted Intervention--MICCAI 2019: 22nd International Conference, Shenzhen, China, October 13--17, 2019, Proceedings, Part VI 22}, pages 175--184. Springer, 2019{\natexlab{b}}.

\bibitem[Wang et~al.(2022)Wang, Wu, Agarwal, and Sun]{wang2022medclip}
Zifeng Wang, Zhenbang Wu, Dinesh Agarwal, and Jimeng Sun.
\newblock Medclip: Contrastive learning from unpaired medical images and text.
\newblock \emph{arXiv preprint arXiv:2210.10163}, 2022.

\bibitem[Xie et~al.(2023)Xie, Yang, Guan, Zhang, Wu, and Xia]{xie2023attention}
Yutong Xie, Bing Yang, Qingbiao Guan, Jianpeng Zhang, Qi~Wu, and Yong Xia.
\newblock Attention mechanisms in medical image segmentation: A survey.
\newblock \emph{arXiv preprint arXiv:2305.17937}, 2023.

\bibitem[Xu et~al.(2020)Xu, Jin, Wang, and Huang]{xu2020multi}
Haotian Xu, Xiaobo Jin, Qiufeng Wang, and Kaizhu Huang.
\newblock Multi-scale attention consistency for multi-label image classification.
\newblock In \emph{Neural Information Processing: 27th International Conference, ICONIP 2020, Bangkok, Thailand, November 18--22, 2020, Proceedings, Part IV 27}, pages 815--823. Springer, 2020.

\bibitem[Ye and Liu(2012)]{ye2012sparse}
Jieping Ye and Jun Liu.
\newblock Sparse methods for biomedical data.
\newblock \emph{ACM Sigkdd Explorations Newsletter}, 14\penalty0 (1):\penalty0 4--15, 2012.

\bibitem[Yu and Koltun(2016)]{dilated}
Fisher Yu and Vladlen Koltun.
\newblock Multi-scale context aggregation by dilated convolutions.
\newblock In Yoshua Bengio and Yann LeCun, editors, \emph{4th International Conference on Learning Representations, {ICLR} 2016, San Juan, Puerto Rico, May 2-4, 2016, Conference Track Proceedings}, 2016.
\newblock URL \url{http://arxiv.org/abs/1511.07122}.

\bibitem[Zeiler(2014)]{zeiler2014visualizing}
MD~Zeiler.
\newblock Visualizing and understanding convolutional networks.
\newblock In \emph{European conference on computer vision/arXiv}, volume 1311, 2014.

\bibitem[Zhou et~al.(2016)Zhou, Khosla, Lapedriza, Oliva, and Torralba]{zhou2016learning}
Bolei Zhou, Aditya Khosla, Agata Lapedriza, Aude Oliva, and Antonio Torralba.
\newblock Learning deep features for discriminative localization.
\newblock In \emph{Proceedings of the IEEE conference on computer vision and pattern recognition}, pages 2921--2929, 2016.

\end{thebibliography}


\clearpage
\appendix
\section{Ablation Experiments}

\subsection{Removing spatial attention}
\label{remove_spatial}

\begin{table*}[!htb]\centering
\caption{Ablation study for deciding how to compute cross-attention. Results are based on 5 trial averages.}
\resizebox{\textwidth}{!}{%
\begin{tabular}{@{}ccccccc@{}}\toprule
& Attention Computation &
  Avg. Test Acc. $\pm$ Std. &
  Avg. Test Spec. $\pm$ Std. &
  Avg. Test Sens. $\pm$ Std. &
  Avg. Test AUROC $\pm$ Std. &
  Avg. Test F1 Score $\pm$ Std. \\ \midrule
\multirow{3}{*}{\rotatebox[origin=c]{90}{\large\textbf{Topcon}}}
& Only Channel Cross-attention &
    \textbf{0.8183 ± 0.0340} &
    \textbf{0.8290 ± 0.0604} &
    \textbf{0.8063 ± 0.0203} &
    \textbf{0.8176 ± 0.0339} &
    \textbf{0.8204 ± 0.0288} \\
& Spatial + Channel Cross-attention &
    0.8018 ± 0.0206 &
    0.8176 ± 0.0648 &
    0.7848 ± 0.0558 &
    0.8012 ± 0.0211 &
    0.8018 ± 0.0171 \\
& EPA \citep{unetrpp}&
    0.7872 ± 0.0404 &
    0.8235 ± 0.0372 &
0.7539 ± 0.0593 &
    0.7887 ± 0.0385 &
    0.7831 ± 0.0429 \\ \bottomrule
\end{tabular}%
}
\label{tb6}
\end{table*}

Efficient Paired Attention (EPA), introduced by \cite{unetrpp}, was designed to efficiently compute spatial and channel self-attention for 3D feature volumes. However, its reliance on projections to condense spatial dimensions into smaller vectors introduces a significant bottleneck and increases the parameter count. In our cross-attention method, we address this limitation by computing attention exclusively along the channel dimension. As demonstrated in Table \ref{tb6}, incorporating cross-attention across both spatial and channel dimensions provides no tangible benefit beyond an increased parameter count.

\subsection{Selecting $\lambda$ for Multi-Task Fine-Tuning}
\label{lambda_choice}

\begin{table*}[!htb]\centering
\caption{A comparison for selecting which $\lambda$ to utilize for each dataset. These results are based on 3 trial averages.}
\resizebox{\textwidth}{!}{%
\begin{tabular}{@{}ccccccc@{}}\toprule
& $\lambda$ Value &
  Avg. Test Acc. $\pm$ Std. &
  Avg. Test Spec. $\pm$ Std. &
  Avg. Test Sens. $\pm$ Std. &
  Avg. Test AUROC $\pm$ Std. &
  Avg. Test F1 Score $\pm$ Std. \\ \midrule
\multirow{5}{*}{\rotatebox[origin=c]{90}{\large\textbf{Topcon}}}
& 0.25 &
    0.8073 ± 0.0198 &
    0.7730 ± 0.0469 &
    0.8360 ± 0.0173 &
    0.8045 ± 0.0231 &
    \textbf{0.8197 ± 0.0109} \\
& 0.50 &
    0.7859 ± 0.0114 &
    0.7687 ± 0.0177 &
    0.8003 ± 0.0173 &
    0.7845 ± 0.0111 &
    0.7957 ± 0.0182 \\
& 0.75 &
    \textbf{0.8104 ± 0.0086} &
    \textbf{0.8209 ± 0.0606} &
    0.8011 ± 0.0412 &
    \textbf{0.8110 ± 0.0104} &
    0.8150 ± 0.0097 \\
& 0.90 &
    0.7951 ± 0.0043 &
    0.7429 ± 0.0646 &
    \textbf{0.8420 ± 0.0636} &
    0.7924 ± 0.0056 &
    0.8103 ± 0.0136 \\ 
& 1.0 &
    0.7951 ± 0.0043 &
    0.7429 ± 0.0646 &
    0.8420 ± 0.0636 &
    0.7924 ± 0.0056 &
    0.8103 ± 0.0136 \\ 
\bottomrule
\multirow{5}{*}{\rotatebox[origin=c]{90}{\large\textbf{Zeiss}}}
& 0.25 &
    0.8617 ± 0.0157 &
    \textbf{0.8987 ± 0.0588} &
    0.8240 ± 0.0567 &
    0.8613 ± 0.0161 &
    0.8496 ± 0.0140 \\
& 0.50 &
    \textbf{0.8681 ± 0.0115} &
    0.8625 ± 0.0207 &
    \textbf{0.8761 ± 0.0461} &
    \textbf{0.8693 ± 0.0129} &
    \textbf{0.8627 ± 0.0141} \\
& 0.75 &
    0.8519 ± 0.0183 &
    0.8634 ± 0.0647 &
    0.8426 ± 0.0377 &
    0.8530 ± 0.0153 &
    0.8439 ± 0.0160 \\
& 0.90 &
    0.8680 ± 0.0070 &
    0.8926 ± 0.0276 &
    0.8428 ± 0.0363 &
    0.8677 ± 0.0080 &
    0.8582 ± 0.0073 \\ 
& 1.0 &
    0.7332 ± 0.0340 &
    0.8801 ± 0.0749 &
    0.5700 ± 0.0291 &
    0.7250 ± 0.0295 &
    0.6707 ± 0.0188 \\ 
\bottomrule
\end{tabular}
}
\label{tb4}
\end{table*}

In our multi-task fine-tuning framework, the overall loss function is defined as:
\begin{equation}
L_{\text{multi-task}} = (1-\lambda) L_{\text{supervised}} + \lambda L_{\text{unsupervised}}.
\end{equation}
The parameter $\lambda$ controls the balance between the supervised loss $L_{\text{supervised}}$ and the unsupervised loss $L_{\text{unsupervised}}$. A higher $\lambda$ increases the influence of unsupervised learning, while a lower $\lambda$ prioritizes supervised learning.To determine the optimal $\lambda$ value, we conducted ablation studies on two datasets: Topcon and Zeiss. The results, shown in Table \ref{tb4}, indicate that the impact of $\lambda$ varies across datasets.

For the Topcon dataset, we observe that $\lambda=0.75$ achieves the highest test accuracy (0.8104), specificity (0.8209), and AUROC (0.8110). The F1 score (0.8150) is also close to the highest value obtained. This suggests that emphasizing the unsupervised component at this weight contributes to improved model performance, likely by leveraging additional structure in the data to refine decision boundaries. For the Zeiss dataset, $\lambda=0.50$ performs best, yielding the highest test accuracy (0.8681), sensitivity (0.8761), AUROC (0.8693), and F1 score (0.8627). Interestingly, while $\lambda=0.90$ achieves high specificity (0.8926), it does not lead to better performance across the other metrics. This indicates that the optimal $\lambda$ value depends on dataset characteristics, and a moderate balance between supervised and unsupervised learning is preferable to avoid overfitting \citep{mirzazadeh2023atcon} to patterns that do not generalize well.

Across both datasets, we observe that relying solely on the unsupervised loss for fine-tuning ($\lambda = 1.0$) leads to model degeneration. This suggests that excessive dependence on unsupervised learning can degrade performance by overemphasizing features that do not align well with the primary classification task, ultimately reducing the model’s effectiveness.

\subsection{Selecting the Unsupervised Loss Function}
\label{loss_choice}

\begin{table*}[!htb]\centering
\caption{Ablation study for selecting the loss function with $\lambda = 0.75$. Results are based on 3 trial averages.}
\resizebox{\textwidth}{!}{%
\begin{tabular}{@{}ccccccc@{}}\toprule
& Loss Function &
  Avg. Test Acc. $\pm$ Std. &
  Avg. Test Spec. $\pm$ Std. &
  Avg. Test Sens. $\pm$ Std. &
  Avg. Test AUROC $\pm$ Std. &
  Avg. Test F1 Score $\pm$ Std. \\ \midrule
\multirow{4}{*}{\rotatebox[origin=c]{90}{\large\textbf{Topcon}}}
& MSE &
    \textbf{0.8104 ± 0.0086} &
    \textbf{0.8209 ± 0.0606} &
    0.8011 ± 0.0412 &
    \textbf{0.8110 ± 0.0104} &
    \textbf{0.8150 ± 0.0097} \\
& SSIM &
    0.7737 ± 0.0189 &
    0.7826 ± 0.0134 &
    0.7676 ± 0.0344 &
    0.7751 ± 0.0172 &
    0.7798 ± 0.0169 \\
& Pearson &
    0.7920 ± 0.0114 &
    0.8014 ± 0.0198 &
    0.7842 ± 0.0390 &
    0.7928 ± 0.0099 &
    0.7971 ± 0.0151 \\
& Gaussian Pearson &
    0.7523 ± 0.0899 &
    0.6724 ± 0.1843 &
\textbf{0.8332 ± 0.0559} &
    0.7528 ± 0.0889 &
    0.7824 ± 0.0658 \\ \bottomrule
\end{tabular}%
}
\label{tb5}
\end{table*}

Given that $\lambda=0.75$ provided strong performance in the Topcon dataset, we further examined the effect of different loss functions for $L_{\text{unsupervised}}$ at this setting. Table \ref{tb5} presents the results of this ablation study.

Among the four loss functions tested (MSE, SSIM, Pearson, and Gaussian Pearson), MSE consistently outperformed the others in terms of accuracy (0.8104), specificity (0.8209), AUROC (0.8110), and F1 score (0.8150). This suggests that minimizing mean squared error in the unsupervised loss effectively preserves useful feature representations while avoiding excessive penalization of small variations in data distributions.

SSIM and Pearson correlation loss yield suboptimal performance compared to MSE. SSIM, designed for structural similarity, performs worse across all metrics, indicating that it may not sufficiently preserve relevant feature distributions in the context of medical imaging classification. Pearson correlation, while improving over SSIM, does not reach the accuracy or F1-score achieved with MSE, possibly due to its focus on linear relationships rather than absolute differences.

Interestingly, Gaussian Pearson loss achieves the highest sensitivity (0.8332), indicating strong recall for positive cases. However, its low specificity (0.6724) and relatively lower AUROC (0.7528) suggest an overemphasis on certain patterns that may not generalize well. This highlights a trade-off when using loss functions that heavily favor recall over precision.

Based on these findings, we conclude that MSE is the most effective choice for $L_{\text{unsupervised}}$ in our multi-task fine-tuning framework, providing a balance between sensitivity and specificity while maximizing overall performance. Future work may explore hybrid loss formulations to further optimize model generalization across datasets.

\subsection{Sampling Strategy}
\label{sampling_strategy}

\begin{table*}[!htb]\centering
\caption{Ablation study for deciding on sampling strategy. Results are based on 5 trial averages.}
\resizebox{\textwidth}{!}{%
\begin{tabular}{@{}ccccccc@{}}\toprule
& Data Sampling Method &
  Avg. Test Acc. $\pm$ Std. &
  Avg. Test Spec. $\pm$ Std. &
  Avg. Test Sens. $\pm$ Std. &
  Avg. Test AUROC $\pm$ Std. &
  Avg. Test F1 Score $\pm$ Std. \\ \midrule
\multirow{3}{*}{\rotatebox[origin=c]{90}{\large\textbf{Topcon}}}

& No Sampling &
    \textbf{0.9500 ± 0.0186} &
    \textbf{0.9636 ± 0.0204} &
    0.7139 ± 0.0283 &
    \textbf{0.8388 ± 0.0143} &
    0.6214 ± 0.1005 \\
& Weighted Loss &
    0.9299 ± 0.0260 &
    0.9409 ± 0.0263 &
    0.7368 ± 0.0177 &
    \textbf{0.8388 ± 0.0172} &
    0.5509 ± 0.0819 \\ 
& Random Undersampling &
    0.8183 ± 0.0340 &
    0.8290 ± 0.0604 &
    \textbf{0.8063 ± 0.0203} &
    0.8176 ± 0.0339 &
    \textbf{0.8204 ± 0.0288} \\ \midrule
\multirow{3}{*}{\rotatebox[origin=c]{90}{\large\textbf{Zeiss}}}
& No Sampling &
    \textbf{0.8753 ± 0.0207} &
    0.7313 ± 0.0643 &
    \textbf{0.9245 ± 0.0266} &
    0.8279 ± 0.0299 &
    \textbf{0.9178 ± 0.0139} \\
& Weighted Loss &
    0.8723 ± 0.0206 &
    0.7237 ± 0.0653 &
    0.9225 ± 0.0298 &
    0.8231 ± 0.0262 &
    0.9155 ± 0.0156 \\
& Random Undersampling &
    0.8315 ± 0.0105 & 
    \textbf{0.8246 ± 0.0336} & 
    0.8425 ± 0.0388 & 
    \textbf{0.8336 ± 0.0130} & 
    0.8311 ± 0.0134 \\
    \bottomrule
\end{tabular}%
}
\label{sampling_table}
\end{table*}

\begin{table*}[!htb]
\centering
\caption{Federated 3D CNN Performance on Topcon and Zeiss Test Sets (AUROC and F1-Score, Mean ± Std)}
\resizebox{\textwidth}{!}{%
\begin{tabular}{lcccc}
\toprule
\textbf{Data Used} & \textbf{Topcon AUROC} & \textbf{Topcon F1} & \textbf{Zeiss AUROC} & \textbf{Zeiss F1} \\
\midrule
Original Topcon & 0.7658 ± 0.0476 & 0.7829 ± 0.0335 & N/A & N/A \\
Original Zeiss & N/A & N/A & \textbf{0.8438 ± 0.0495} & \textbf{0.8439 ± 0.0392} \\
Original Topcon + Original Zeiss & \textbf{0.7875 ± 0.0230} & \textbf{0.8061 ± 0.0225} & 0.8382 ± 0.0436 & 0.8311 ± 0.0374 \\
\bottomrule
\end{tabular}
}
\label{tab:federated_results}
\end{table*}

Since Dataset 1 (Topcon) contains 4932 non-glaucomatous and 272 glaucomatous samples, and Dataset 2 (Zeiss) contains 263 non-glaucomatous and 847 glaucomatous samples, we applied resampling strategies to mitigate class imbalance and prevent the model from overfitting to the majority class. 

The first method we used was a class-weighted version of the binary cross-entropy (BCE) loss for supervised training. The weights were computed separately for each dataset based on the proportion of glaucomatous ($y_x = 1$) and non-glaucomatous ($y_x = 0$) samples. Let the predicted glaucoma probability be $P_x \in [0, 1]$, and the ground truth label be $y_x \in \{0, 1\}$. The class-weighted BCE loss is defined as:

\begin{equation}
    {L}_{\text{weighted}} = -\frac{1}{N} \sum_{i=1}^{N} \left[ w_1 y_x \log P_x + w_0 (1 - y_x) \log (1 - P_x) \right]
    \label{eq:weighted_bce}
\end{equation}

To emphasize the contribution of the minority class, we used the following weights:
\[
w_0^{(1)} = \frac{272}{4932 + 272}, \quad w_1^{(1)} = 1 - w_0^{(1)} \quad \text{(Topcon)}
\]
\[
w_0^{(2)} = \frac{847}{847 + 263}, \quad w_1^{(2)} = 1 - w_0^{(2)} \quad \text{(Zeiss)}
\]

These weights were applied per sample based on the ground truth label. This ensures that glaucomatous or non-glaucomatous cases that are underrepresented have a proportionately greater influence during optimization.

The second method involved random undersampling of the majority class to construct a balanced dataset during training. For each training trial, we randomly sampled an equal number of glaucomatous and non-glaucomatous volumes to form a balanced training batch. This approach guarantees equal representation of both classes during each optimization step. While it reduces the total number of available samples, it helps prevent the model from becoming biased toward the majority class and often improves performance on the minority class.

In Table~\ref{sampling_table}, we report the results of three sampling strategies: No Sampling, Weighted Loss, and Random Undersampling, applied during training of the base $\text{AI-CNet3D}_{H}$ model in Step 1 (Figure~\ref{fig:full-model}). These results show that while No Sampling and Weighted Loss improve some metrics, they fall short in others. Sensitivity is low when training on Topcon data, which has more negative examples. Conversely, specificity is low for Zeiss data, which contains more positive examples. Random undersampling results in the most balanced performance across all metrics, suggesting that neither class dominates the learning process.

\subsection{Federated Training}
\label{federated_training}

We also evaluate a scenario where the model is trained using combined data from both datasets. Since Dataset 1 (Topcon) contains both macula and ONH regions while Dataset 2 (Zeiss) contains only ONH, we standardize the data by cropping Topcon volumes to include only the ONH region and resizing them to 64×128×64 to match the Zeiss volume dimensions.

Table \ref{tab:federated_results} presents the performance comparison between the base 3D CNN trained on individual datasets versus a combined approach using the FedAvg protocol \citep{mcmahan2017communication}. The results indicate that combining datasets does not yield performance improvements when evaluated on their respective test sets. Notably, the combined training approach shows modest improvements on the Topcon test set but slightly reduced performance on the Zeiss test set compared to training exclusively on Zeiss data.

This lack of substantial improvement can be attributed to the fundamental differences between the two OCT acquisition systems, which produce volumes with distinct noise characteristics and imaging artifacts. Future research will focus on domain adaptation techniques to harmonize these different volume types and achieve more consistent OCT data representation across platforms.

\end{document}